\documentclass[11pt,a4paper]{article}%

\usepackage[utf8]{inputenc}
\usepackage[T1]{fontenc}\usepackage{lmodern}
\input{glyphtounicode}\pdfgentounicode=1
\usepackage[numbers,sort&compress]{natbib}
\usepackage{authblk}
\usepackage[a4paper,margin=2cm]{geometry}

\usepackage{graphicx}%
\usepackage{multirow}%
\usepackage{amsmath,amssymb,amsfonts}%
\makeatletter
\let\GFunkSavedEqref\eqref%
\makeatother
\usepackage{amsthm}%
\usepackage{mathrsfs}
\DeclareFontFamily{U}{rsfs}{\skewchar\font127}
\DeclareFontShape{U}{rsfs}{m}{n}{<-6.5> rsfs5 <6.5-8> rsfs7 <8-> rsfs10}{}%
\usepackage[title]{appendix}%
\usepackage{xcolor}%
\usepackage{textcomp}%
\usepackage{manyfoot}%
\usepackage{booktabs}\usepackage{tabularx}%
\usepackage{algorithm}%
\usepackage{algorithmicx}%
\usepackage{algpseudocode}%
\usepackage{listings}%
\usepackage{float}

\usepackage{amsmath,amsfonts,bm}

\def\eqref#1{equation~\ref{#1}}

\def\1{\bm{1}}

\DeclareMathAlphabet{\mathsfit}{\encodingdefault}{\sfdefault}{m}{sl}
\SetMathAlphabet{\mathsfit}{bold}{\encodingdefault}{\sfdefault}{bx}{n}

\makeatletter
\let\eqref\GFunkSavedEqref%
\makeatother

\usepackage{url}
\usepackage{siunitx}
\usepackage{nicefrac}%
\usepackage{microtype}%
\usepackage{soul}

\usepackage{subcaption}

\usepackage{diagbox}

\usepackage[hidelinks]{hyperref}
\hypersetup{pdftitle={Graph Fourier Neural Kernels: Learning Full PDE Trajectories Across Geometries and Anisotropic Parameter Fields},pdfauthor={Shane E. Loeffler et al.}}
\usepackage{placeins}%
\theoremstyle{plain}%

\theoremstyle{definition}%

\theoremstyle{remark}%

\title{Graph Fourier Neural Kernels: Learning Full PDE Trajectories Across Geometries and Anisotropic Parameter Fields}

\author[1]{Shane~E.~Loeffler}
\author[1,2]{Zan~Ahmad}
\author[1,2]{Shiyi~Chen}
\author[1]{Syed~Yusuf~Ali}
\author[1]{Carolyna~Yamamoto}
\author[1]{Kensuke~Sakata}
\author[1,2]{Dan~M.~Popescu}
\author[1]{Alana~Yee}
\author[1]{Yash~Lal}
\author[1,2,*,\textdagger]{Natalia~Trayanova}
\author[2,3,\textdagger]{Mauro~Maggioni}

\affil[1]{Department of Biomedical Engineering, Johns Hopkins University, Baltimore, MD, USA}
\affil[2]{Department of Applied Mathematics and Statistics, Johns Hopkins University, Baltimore, MD, USA}
\affil[3]{Department of Mathematics, Johns Hopkins University, Baltimore, MD, USA}

\date{}

\begin{document}
\maketitle

{\centering\small
$^{*}$Corresponding author. Email: \texttt{ntrayanova@jhu.edu} (N.T.). \par
Also: \texttt{mauromaggionijhu@icloud.com}, \texttt{sloeffl2@jhu.edu}, \texttt{zahmad6@jhu.edu}.\par
$^{\dagger}$ These authors are equally contributing senior co-authors.\par}
\vspace{1em}

\begin{abstract}
\noindent
High-fidelity simulations of time-dependent nonlinear partial differential equations (PDEs) are expensive to repeat across families of domains and parameter fields, a cost that recurs in design optimization, uncertainty quantification, and computational medicine. Most neural-operator surrogates require each domain to be mapped onto a common grid or reference shape, and few predict the full time course. We introduce Graph Fourier Neural Kernels (G-FuNK), a neural operator for second-order semilinear PDEs with a diffusive principal part that transfers across geometries and anisotropic diffusion tensors without coordinate maps or diffeomorphisms. A weighted graph and its Laplacian supply a spectral basis that encodes geometry and anisotropy. Shared layers act in this basis, at a cost linear in mesh size after one partial eigendecomposition per geometry, and a neural ODE integrator propagates the learned generator to full spatiotemporal trajectories on unseen domains. Across anisotropic heat flow, stiff reaction-diffusion, FitzHugh-Nagumo, and patient-specific left-atrial electrophysiology on CT-derived surfaces, G-FuNK reproduces held-out trajectories with relative $L^2$ error between 0.046 and 0.160. Its accuracy is comparable to that of a geometry-aware Fourier neural operator on regular domains, and its predictions are unchanged when a rectangular domain is rotated. On left atria, whose venous and valvular openings prevent mapping onto a reference shape, it predicts an 85-ms trajectory in under half a second, against about 5 minutes for finite elements, with a mean activation-time error of 1.73 ms. Accuracy depends on graph construction and is highest when the learned generator is integrated at its training time step.
\end{abstract}

\section*{Significance Statement}
\noindent
Many problems in engineering and medicine require solving the same physical equations repeatedly on differently shaped objects, such as the hearts of different patients. Each high-fidelity simulation takes minutes or longer, which limits how many scenarios can be explored. Machine-learning models trained on simulations run much faster, but most assume that every shape can be mapped onto a common grid. That assumption fails for structures with holes, such as the left atrium of the heart with its vein and valve openings. We describe a model that builds the shape and the direction-dependent tissue properties into the functions that represent the solution. It predicts how electrical activation spreads across patient-specific atria in under a second, against minutes for a standard solver.

\noindent\textbf{Keywords:} Neural operators; Spectral learning; Geometric deep learning; Partial differential equations

\vspace{0.5em}

\section{Introduction}

Many physical, biological, and engineering systems are governed by partial differential equations whose dynamics depend jointly on the geometry of the domain, on spatially varying and often strongly anisotropic material parameters, and on boundary conditions. The dependence is strong in computational medicine, where the same equations must be solved for many patient-specific anatomies, each with its own anisotropic parameter field. In cardiac electrophysiology this field is the myocardial fiber field, which sets the local conduction tensor. Each high-fidelity finite-element solve is expensive, and the cost is incurred again for every new geometry, parameter field, and initial condition. A surrogate model for this setting needs four properties. It must transfer across a family of domains rather than a single fixed one. It must adapt to spatially varying anisotropic diffusion tensors. It must produce full spatiotemporal trajectories rather than a fixed-time snapshot or a single derived scalar. And it must scale to the mesh sizes of clinical anatomies. To our knowledge, no existing neural-operator framework has all four.

Each of the main existing architectures lacks at least one. Spectral methods on regular domains, namely the Fourier Neural Operator (FNO) \cite{li2020fourier} and its geometry-aware variants Geo-FNO \cite{li2023fourier}  and GINO \cite{li2024geometry}, rely on a global Euclidean grid amenable to the FFT; they extend to irregular inputs by learning a deformation onto a reference grid, a map that is coordinate-dependent and that assumes each domain can be smoothly deformed onto a canonical one. Domain-mapping operators such as DIMON \cite{yin2024dimon,ahmad2025dlno}, an extension of DeepONet-based operators \cite{lu2021learning, jin2022mionet}, transfer across diffeomorphic families by computing explicit diffeomorphisms, at a cost that scales superlinearly in mesh size. Message-passing graph neural operators \cite{gilmer2020message,kipf2016semi,li2020neural,simonovsky2017dynamic,iakovlev2020learning,li2020multipole} aggregate information locally, which can limit how well they represent the long-range directional coupling imposed by anisotropic diffusion. Spectral graph constructions provide the global mixing that local aggregation lacks \cite{defferrard2016convolutional,bo2023survey,wang2022powerful,yang2022new}, and a separate line of work targets long-horizon trajectory prediction \cite{chen2023implicit,zhang2024sinenet,he2024sequential,regazzoni2024learning}, but neither addresses varying geometry and varying anisotropic parameter fields together. Section~\ref{sec12} gives a fuller account.

For atrial anatomy, some of these methods cannot be applied at all. The human left atrium, with its five venous and valvular orifices, cannot be diffeomorphically mapped to a cube, torus, or sphere, and embedding it in a bounding box enlarges the mesh and introduces restriction artifacts that are especially harmful to Fourier-based reconstruction.

Here we introduce Graph Fourier Neural Kernels (G-FuNK). In G-FuNK, geometry and anisotropy are built into the basis in which the operator is learned, instead of being supplied as inputs in a fixed coordinate system. For each domain and diffusion tensor, we build a weighted graph whose Laplacian approximates the leading elliptic term of the PDE, with edge weights set by the local diffusion tensor. The eigenvectors of this Laplacian form a generalized Fourier basis that encodes the geometry and the anisotropy. Learned layers act in this basis. Because the basis is computed in advance for each instance and the weights are shared, the model transfers across discretizations, geometries, and parameter fields without retraining or coordinate maps. A neural ODE integrator \cite{chen2018neural} then integrates the learned generator to full trajectories from new initial conditions on unseen domains. Because the construction needs no diffeomorphisms and no global parameterization, its cost is linear in mesh size, unlike the superlinear cost of domain-mapping approaches.

\noindent\textbf{Contribution.}
G-FuNK is, to our knowledge, the first neural operator to predict full spatiotemporal trajectories of stiff, strongly anisotropic reaction-diffusion systems across a family of topologically non-trivial patient-specific surfaces, each with its own anisotropic diffusion-tensor field. Because geometry and anisotropy enter through the Laplacian eigenbasis rather than through coordinates, the learned operator does not depend on a global coordinate frame, and it runs on the atrial surface graph without the seam that atrial coordinate systems introduce \cite{zhou2026unifiedframeworkgeometryindependentoperator}. Geo-FNO would require such a coordinate system on the atrium, so we did not run it on the atrial geometries. Inference cost is linear in mesh size, after a one-time graph construction and partial eigendecomposition per geometry (Section~\ref{graph_laplacian_stuff}). On 25 patient-specific left atria, G-FuNK predicts an \SI{85}{\milli\second} trajectory in less than half a second, against an average of 4.99 minutes for the finite-element solver, and the activation times derived from its predictions have errors in the same range as those of operators trained to regress activation time directly \cite{yin2024dimon, Naghavi2026, zhou2026unifiedframeworkgeometryindependentoperator}.

\section{Results}\label{sec2}\label{sec:experiments}
\subsection{Method overview}
We focus on the family of second-order semilinear differential operators, governing the evolution of a function \( u: \Omega_\alpha\times[0,\infty)\rightarrow\mathbb{R} \), $\mathbf{x}\in\Omega_\alpha$, $t\ge0$, in the form:
\begin{equation}
\left\{
\begin{aligned}
    \partial_t u(\mathbf{x},t) &= \underbrace{\nabla_{\mathbf{x}}\cdot
    (\mathbf{K}_{\alpha, \beta}(\mathbf{x}) \nabla_{\mathbf{x}} u(\mathbf{x},t))+\mathcal{S}(\,u(\mathbf{x}, t), \mathbf{x}, \nabla_\mathbf{x} u(\mathbf{x},t)\,)}_{\mathcal{N}(\, u(\mathbf{x}, t), \mathbf{K}_{\alpha, \beta}(\mathbf{x}), \mathbf{x}, t\,)}
    \\ 
    u(\mathbf{x},0) &= u_0(\mathbf{x}) \\
    0 &= \left( \mathbf{K}_{\alpha, \beta}(\mathbf{x}) \nabla u \right) \cdot \mathbf{n}(\mathbf{x}) \:\,,\quad \forall \, \mathbf{x} \: \in \partial \Omega_\alpha\, \\
    \end{aligned}
    \right.
    \label{general_diff_op}
\end{equation}
where $\Omega_\alpha$ is a domain of interest, $\mathbf{K}_{\alpha,\beta}$ is a diffusion tensor field on $\Omega_\alpha$ satisfying uniform ellipticity, and \( \mathcal{N} \) denotes the (true) generator of the dynamics. 

\noindent \textbf{Problem statement}.
We aim to learn a neural operator $\mathcal{N}_{\theta}$ that approximates the generator $\mathcal{N}$, by mapping a current state $u(\cdot,t)$, together with parameters $(\Omega_\alpha,\mathbf{K}_{\alpha, \beta}) \in  \mathcal{M}$, to an approximation of the time derivative $\partial_t u(\cdot,t)$.
We then generate the full solution trajectory $(u_\theta(\mathbf{x},t))_{\mathbf{x}\in\Omega_\alpha, t\in[0,T]}$ by integrating $\mathcal{N}_{\theta}$ using a Neural ODE time-integration framework \cite{chen2018neural}.

\noindent \textbf{Training data}.
The training data consist of trajectories $\{u^{(m)}(x_i,t_\ell)\}_{m=1,i=1,\ell=1}^{M,n_{\alpha^{(m)}},L}$, where the $m$-th trajectory solves \eqref{general_diff_op} on the domain $\Omega_{\alpha^{(m)}}$, with diffusion tensor field $\mathbf{K}_{{\alpha}^{(m)},\beta^{(m)}}$, with some initial condition $u^{(m)}(x_i,0)$, and the $x_i$'s are points scattered in $\Omega_{\alpha^{(m)}}$ (typically, a mesh or point cloud discretizing $\Omega_{\alpha^{(m)}}$).
The times $t_\ell$ are sampled at a timescale that is typically significantly coarser than the time step of the numerical solver used to generate the training data trajectories.
We consider cases with multiple diffusion tensor fields $\mathbf{K}_{{\alpha}^{(m)},\beta^{(m)}}$ on the same domain $\Omega_\alpha$, which corresponds to having $\alpha^{(m)}=\alpha$ for multiple $m$'s, but also cases where for each domain there is a unique diffusion tensor field of interest.

\begin{figure}[!htbp]
    \centering
    \includegraphics[trim={0 0 0 0}, width=\linewidth]{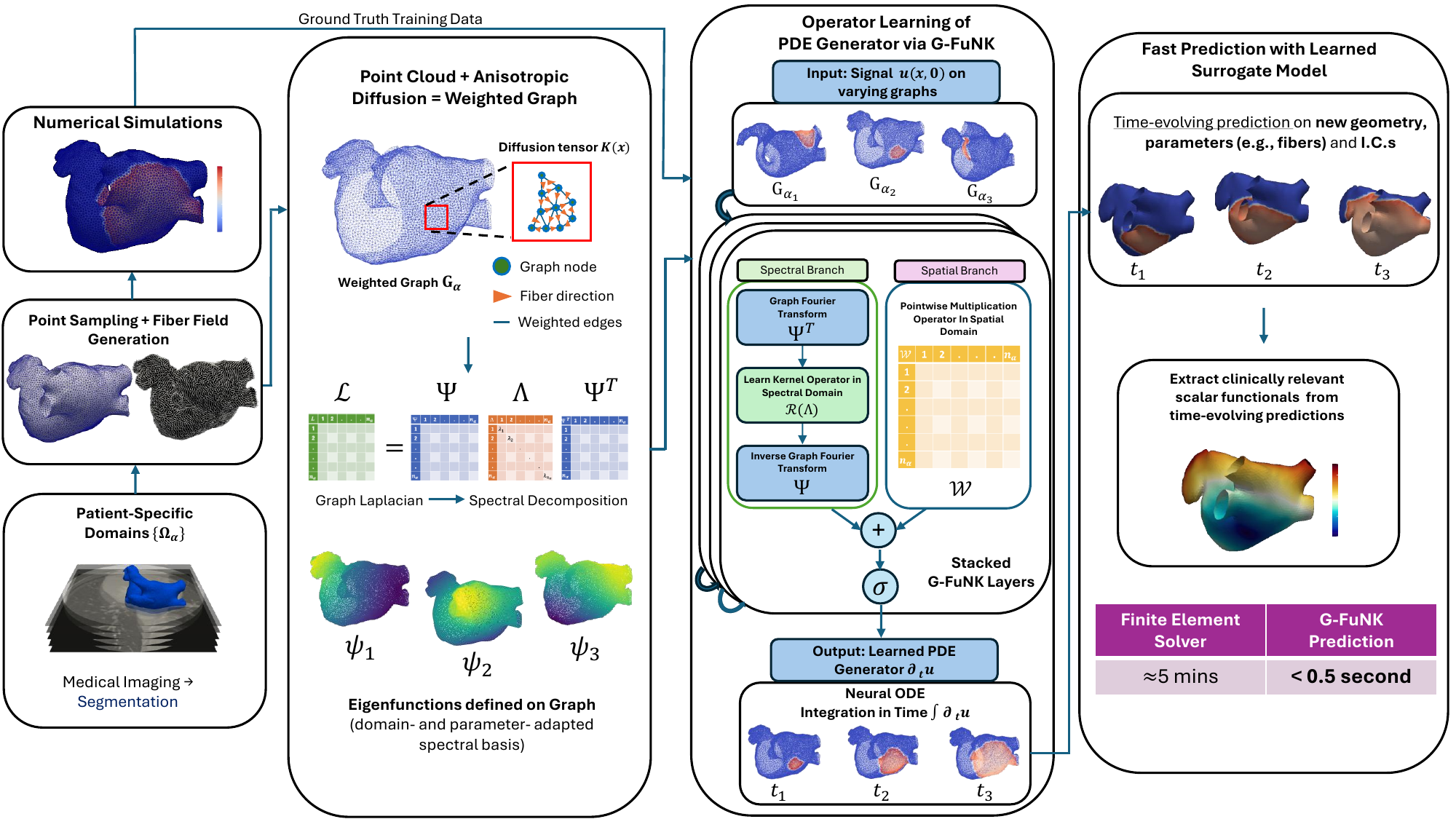}
    \caption{\textbf{Overview of the G-FuNK framework for data-driven cardiac electrophysiology.} Patient-specific anatomies are obtained from medical imaging and processed into computational domains with associated myocardial fiber fields. High-fidelity numerical simulations are used to generate training data. These geometries are then downsampled and converted into graphs, from which graph Laplacians and their spectral decompositions are computed to define domain- and parameter-adapted bases. G-FuNK learns in this spectral domain, combining graph Fourier projections with learned kernel operators to model the temporal evolution of the transmembrane potential. Time integration is performed via a Neural ODE formulation, enabling the prediction of full spatiotemporal trajectories. Once trained, the model provides rapid predictions on previously unseen geometries and parameters, from which clinically relevant functional quantities can be derived. The framework is general and is evaluated across increasingly complex PDE systems, culminating in patient-specific cardiac electrophysiology applications.}
    \label{fig:figure1}
\end{figure}

\noindent \textbf{G-FuNK: an adapted spectral architecture for learning the generator}.
Since the PDEs are semilinear, with the second-order term being a diffusion tensor field, it is natural to leverage a spectral technique, combined with a spatial operator to capture the nonlinear reaction term.
This is very much in the spirit of the FNO framework, which however is restricted to Euclidean domains, where the standard Fourier transform is defined, and grids, amenable to FFTs.
We generalize the FNO framework to general domains and meshes, by mapping the mesh to a weighted graph that incorporates both information about the geometry of the mesh and the domain $\Omega_{\alpha}$, but also, via a tuning of the weights, information about the diffusion tensor field $\mathbf{K}_{\alpha, \beta}$. 
This weighted graph is constructed so that its graph Laplacian approximates the highest order differential term in the PDE. 
The eigenvectors of the graph Laplacian are then used to define a spectral basis for the function space on the graph/discretized domain $\Omega_\alpha$, playing the role of the standard Fourier modes in the FNO framework and its direct variants.
The G-FuNK layer then learns a linear operator in this spectral domain, which is then combined with a spatial operator to capture the nonlinear reaction term.

The steps of this construction are shown schematically in Fig.~\ref{fig:figure1} for the cardiac electrophysiology application and are detailed in section~\ref{graph_laplacian_stuff}.

This construction enables G-FuNK to capture both domain- and parameter-specific structure in its spectral representation, while providing the flexibility and generalization ability required for real-world PDE surrogate modeling across variable geometries and diffusion fields.

\subsection{Examples}
In this section, we evaluate the G-FuNK operator-learning framework across four increasingly challenging problems, beginning with anisotropic diffusion and culminating in patient-specific cardiac electrophysiology simulations. These experiments demonstrate the ability of G-FuNK to learn solution generators for PDEs of the form in (\ref{general_diff_op}), with varying domains, diffusion tensor fields, initial conditions, and other parameters. Details on the construction of the G-FuNK network can be found in section~\ref{graph_laplacian_stuff} and SI Appendix, section~S2.
All benchmarks are organized in the same way: \textit{PDE problem} (governing equations and boundary conditions), \textit{Data preparation} (meshing, coefficients, and reference simulations, citing the Methods sections for reproducibility details), and \textit{G-FuNK performance}, with qualitative figures in each subsection, and errors, together with comparisons to baseline models, aggregated in Table~\ref{table:results}.

\subsubsection{Anisotropic heat equation}
\label{heat}

\paragraph{PDE problem.}
The introduction of anisotropic diffusion is critical in simulating phenomena such as the propagation of heat in materials with fibrous structures, where the diffusive properties vary along different directions.
We consider the heat equation with strong anisotropic diffusion on a 2D unit square $\Omega_{0}$ with Neumann boundary conditions:

\begin{equation}
    \left\{
\begin{aligned}
\partial_t u &= \nabla \cdot (\mathbf{K}_{0, \beta}(\mathbf{x}) \nabla u), \quad \mathbf{x} \in \Omega_{0}, t > 0 \\
0 &= \left( \mathbf{K}_{0, \beta}(\mathbf{x}) \nabla u \right) \cdot \mathbf{n}(\mathbf{x}) \quad \forall \, \mathbf{x} \: \in \: \partial \: \Omega_{0}\,, 
    \label{heat_equation}
\end{aligned}
    \right.
\end{equation}

The diffusion tensor has an anisotropy ratio of $9{:}1$,
\begin{equation}
    \mathbf{K}_{0, \beta}(\mathbf{x}) = 
    \left[
    \begin{array}{cc}
    \hat{F}_{0, \beta}^{(1)}(\mathbf{x}) & \hat{F}_{0, \beta}^{(2)}(\mathbf{x}) \\
    \end{array}
    \right]
    \begin{bmatrix}
    9 & 0 \\
    0 & 1
    \end{bmatrix} 
    \left[
    \begin{array}{cc}
    \hat{F}_{0, \beta}^{(1)}(\mathbf{x}) & \hat{F}_{0, \beta}^{(2)}(\mathbf{x}) \\
    \end{array}
    \right]^T, \label{fiber_eq_1}
\end{equation}
where $\hat{F}_{0, \beta}^{(1)}(\mathbf{x})$ and $\hat{F}_{0, \beta}^{(2)}(\mathbf{x})$ are orthonormal vector fields defining the longitudinal and transverse diffusion directions. In this example, all trajectories are defined on a single domain, while $\beta$ indexes the diffusion field associated with each trajectory $m$.

\paragraph{Data preparation.}
The spatial domain is \emph{fixed} to the unit square for every instance; variability enters only through the tensor field $\mathbf{K}_{0, \beta}(\mathbf{x})$, driven by a random smooth primary direction $\hat{F}^{(1)}_{0, \beta}$ (SI Appendix, Eq.~S1) and an independent random superposition-of-Gaussians initial condition per trajectory.
The learning split uses $500$ training and $100$ test trajectories.
See SI Appendix, section~S1.1, for parameter ranges, meshing, and sampling details.

\paragraph{G-FuNK performance.}
\begin{figure}[!htbp]
    \centering

    \begin{subfigure}{\textwidth}
        \centering
        \includegraphics[trim={10 10 5 5},clip,width=\textwidth]{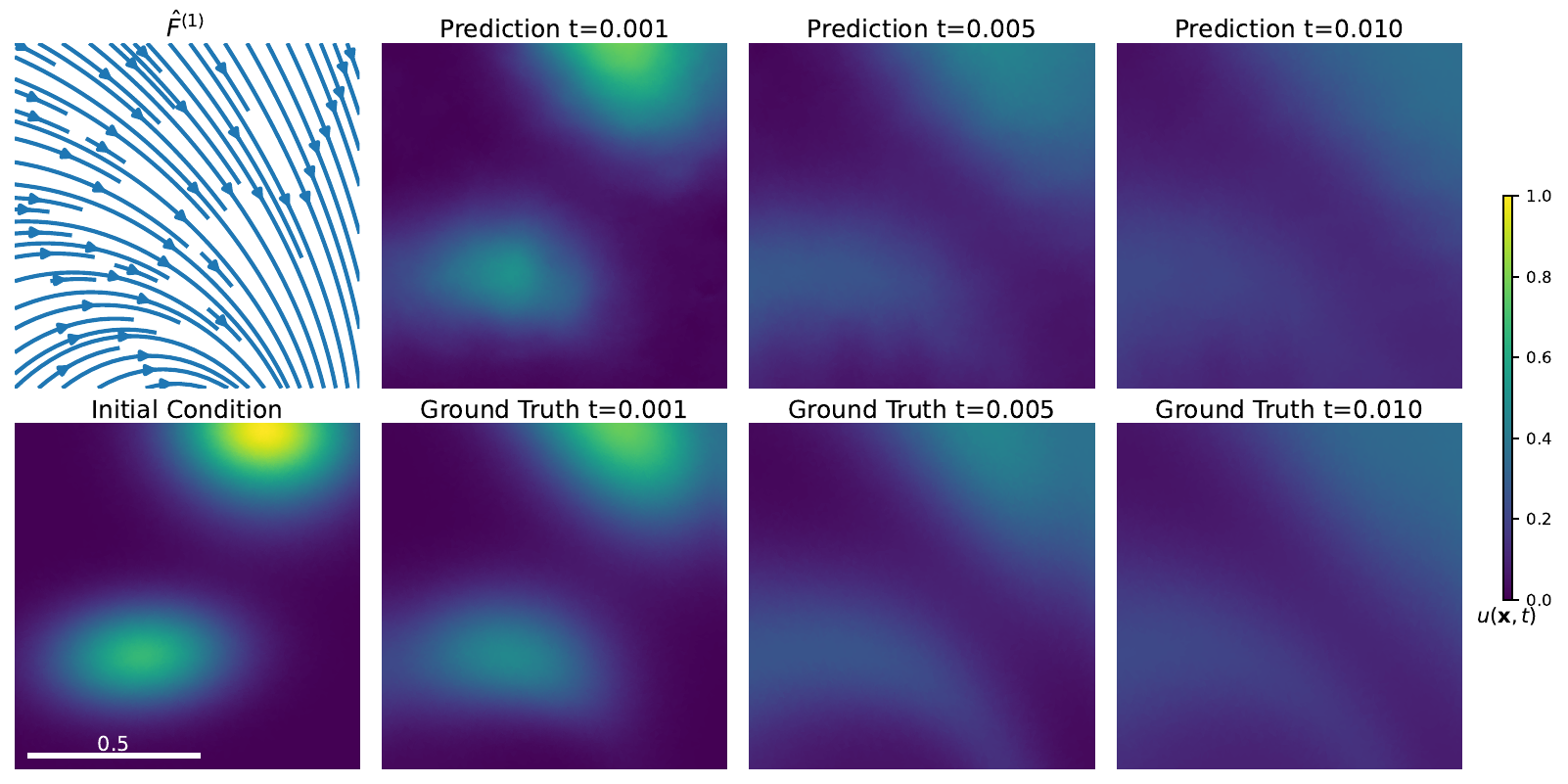}
        \caption{}
        \label{fig:heatA}
    \end{subfigure}

    \vspace{0.5em}
    \hrule
    \vspace{0.5em}

    \begin{subfigure}{\textwidth}
        \centering
        \includegraphics[trim={10 10 5 5},clip,width=\textwidth]{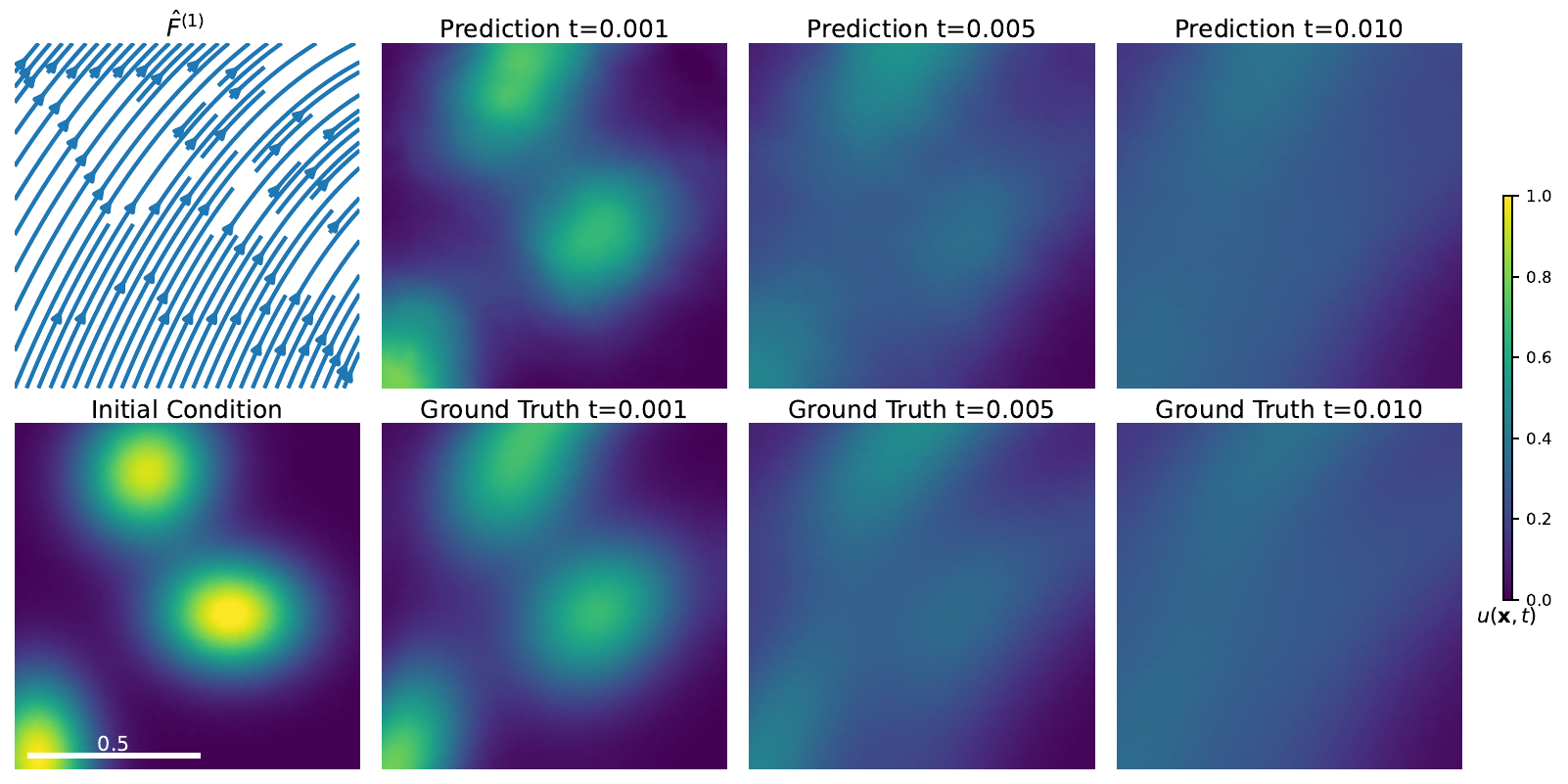}
        \caption{}
        \label{fig:heatB}
    \end{subfigure}

    \caption{\textbf{Heat Equation: G-FuNK vs Target (Two Test Examples).}
        Two representative test examples corresponding to different diffusion fields $\beta$. In each example, the \textit{top-left} panel shows a stream plot of the primary direction of diffusion, $\hat{F}_{0, \beta}^{(1)}$, as defined in SI Appendix, Eq.~S1. The \textit{bottom-left} panel shows the initial condition at $t_0$ with a scale bar of $0.5$ to show the size of the domain. The remaining panels show snapshots of the solution at three later time points, with the G-FuNK predictions shown above the corresponding ground truth.}
    \label{heat_fibs}
\end{figure}
\noindent
Figure~\ref{heat_fibs} shows two representative held-out instances: the stream plots visualize $\hat{F}_{0, \beta}^{(1)}(\mathbf{x})$ that define $\mathbf{K}_{0, \beta}(\mathbf{x})$; prediction snapshots match the numerical ground truth closely (see Table~\ref{table:results}). To examine how accuracy scales with data, we also retrained G-FuNK with 50, 100, and 250 training trajectories; the relative $L^2$ error on the 100 held-out test trajectories decreased monotonically from $0.120$ to $0.099$ to $0.073$, compared with $0.056$ for the full 500-trajectory model (SI Appendix, section~S6).

\subsubsection{2D Nonlinear Reaction--Diffusion}

\paragraph{PDE problem.}
We consider a stiff reaction--diffusion system on a 2D domain that couples the transmembrane potential to myocardial-type ionic dynamics \cite{courtemanche1998ionic}, giving rise to steep, near-discontinuous wavefronts (roughly $-85$ to $20\:\text{mV}$; rates of change exceeding $100\:\text{mV}/\text{ms}$).
An external stimulus current initiates propagation from a random point in the domain for each trajectory.
In terms of $u$, the Courtemanche formulation is
\begin{equation}
    \left\{
\begin{aligned}
\partial_t u &= \nabla \cdot \Sigma \nabla u + \sum_s J_s(u, \mathbf{v}), \quad \mathbf{x} \in \Omega_\alpha, t > 0 \\
\frac{\partial \mathbf{v}}{\partial t} &= \Upsilon(u,\mathbf{v}),\\
0 &= (\Sigma\nabla u) \cdot \mathbf{n}(\mathbf{x}) \quad \forall \, \mathbf{x} \: \in \: \partial \: \Omega_\alpha\,, 
    \label{EP_equation}
\end{aligned}
    \right.
\end{equation}
where $\sum_s J_s$ collects 12 ionic and stimulus currents and $\mathbf{v}\in\mathbb{R}^{20}$ are gating variables \cite{courtemanche1998ionic}. In this example, the anisotropic diffusion is a constant across the various domains $\alpha$, and $\mathbf{K}_{\alpha, \beta}=\Sigma$.

\paragraph{Data preparation.}
Each domain $\Omega_\alpha$ is a \emph{random rectangle}: both edge lengths are drawn independently from $\mathrm{Unif}([15,30])$~mm and meshed for \texttt{openCARP} finite-element runs (Crank--Nicolson, $\Delta t=\SI{5}{\micro\second}$, mean edge length $\SI{0.2}{\milli\meter}$) with $5{:}1$ anisotropy along the horizontal axis.
After simulation, fields are voxel-downsampled ($\SI{0.4}{\milli\meter}$ resolution, average points $3238.94 \pm 885.51$; min-max: 1482-5625) and stored every $\SI{10}{\milli\second}$ up to $\SI{100}{\milli\second}$.
The learning split contains $500$ training and $50$ test trajectories on independently sampled rectangles.
See SI Appendix, section~S1.2, for solver settings and interpolation details.

\begin{figure}[!htbp]
    \centering

    \begin{subfigure}{0.9\textwidth}
        \centering
        \includegraphics[trim={0 0 8 0},clip,width=\textwidth]{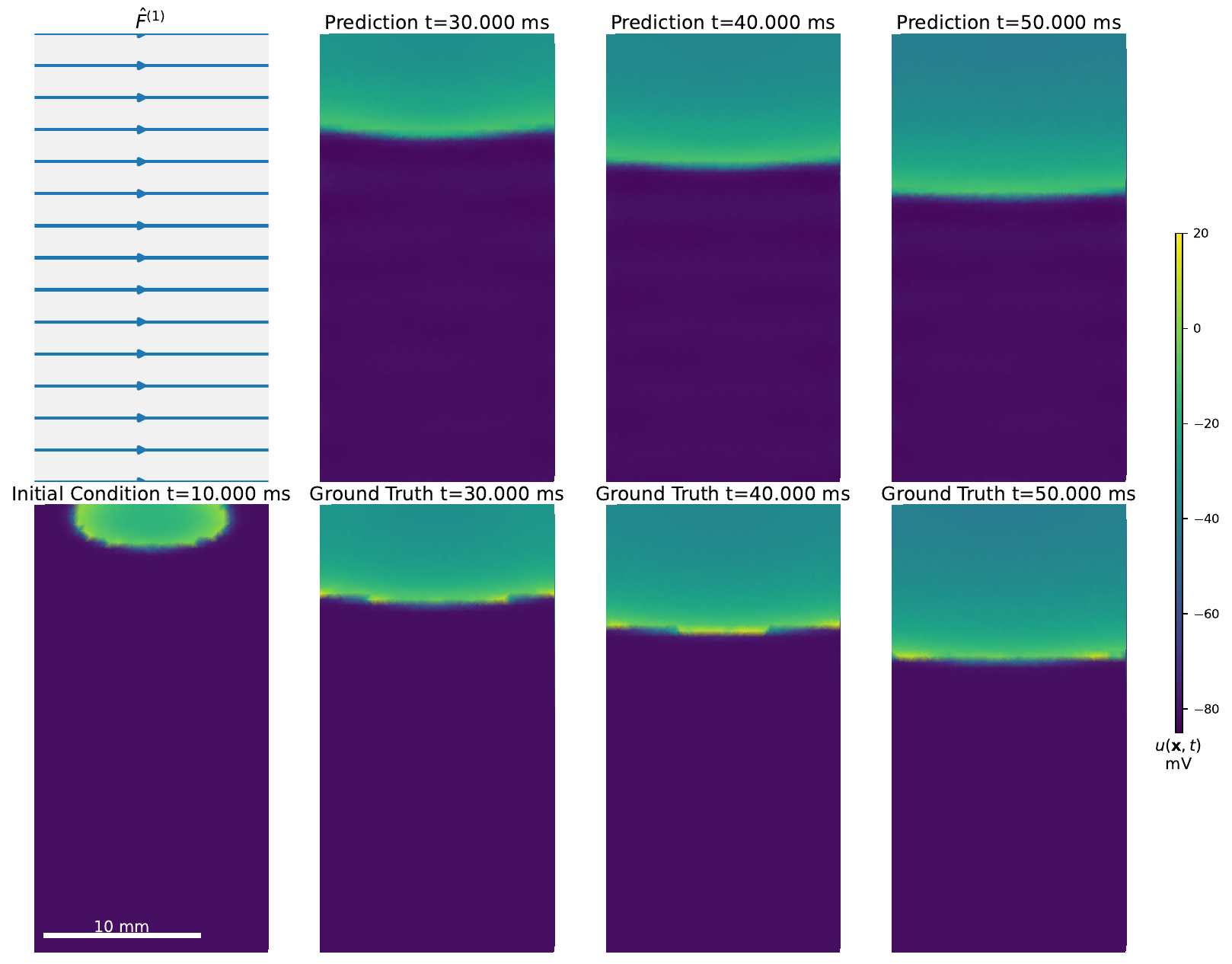}
        \caption{}
        \label{fig:rdA}
    \end{subfigure}

    \vspace{0.5em}
    \hrule
    \vspace{0.5em}

    \begin{subfigure}{0.9\textwidth}
        \centering
        \includegraphics[trim={0 0 8 0},clip,width=\textwidth]{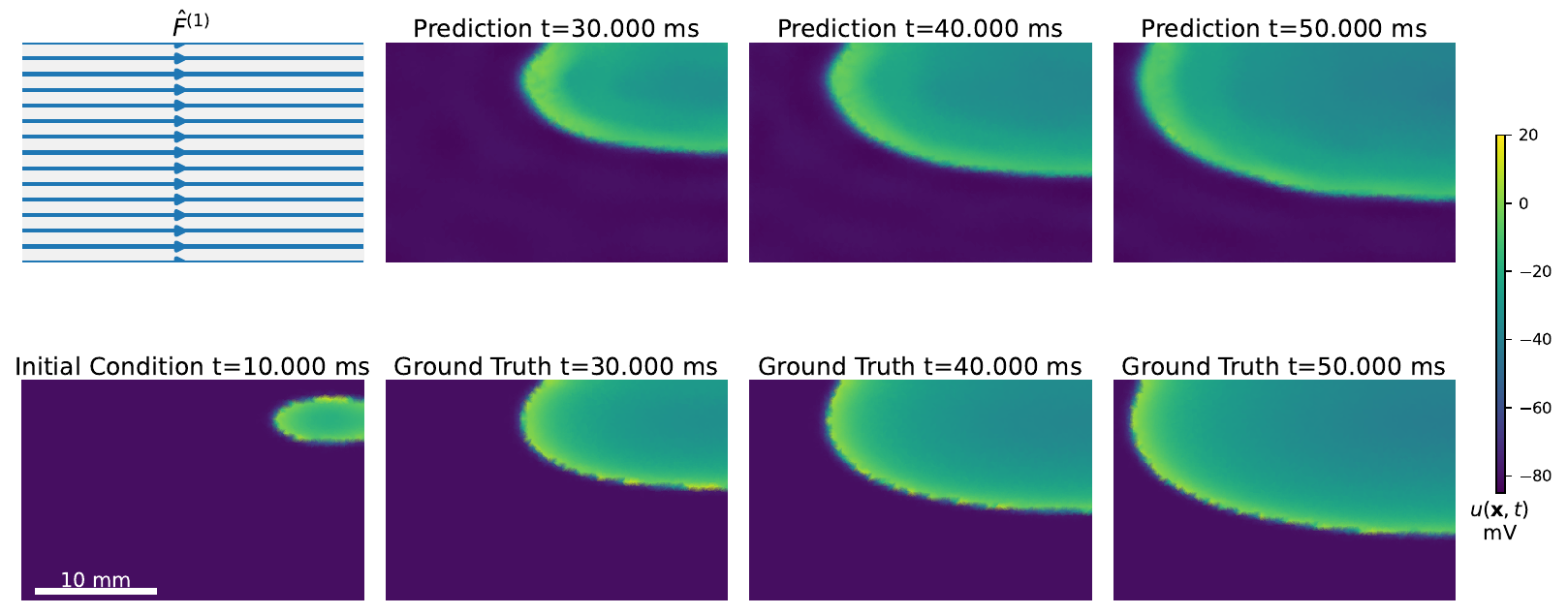}
        \caption{}
        \label{fig:rdB}
    \end{subfigure}

    \caption{\textbf{Reaction--Diffusion on Random Rectangle: G-FuNK vs Target (Two Test Examples).}
        Two representative test examples corresponding to different domains with anisotropic diffusion. In each example, the \textit{top-left} panel shows a stream plot of the primary direction of diffusion (5:1 ratio), $\hat{F}_{\alpha,\beta}^{(1)}$, which is constant for all trajectories in this example. The \textit{bottom-left} panel shows the initial condition at $t_0$ with a scale bar of $\SI{10}{\milli\meter}$ to show the various sizes of the test domains. The remaining panels show snapshots of the solution at three later time points, with the G-FuNK predictions shown above the corresponding ground truth targets.
        }
    \label{sq_rd_fig}
\end{figure}

\paragraph{G-FuNK performance.}
Although~\eqref{EP_equation} couples $21$ state variables overall, we train the surrogate to predict only~$u$, which serves as both input and supervisory signal. 
Figure~\ref{sq_rd_fig} compares G-FuNK to high-fidelity simulations on held-out rectangles. 
\noindent

\subsubsection{FitzHugh--Nagumo}

\paragraph{PDE problem.}
Spatially varying diffusivity shapes dynamics in applications such as arrhythmogenesis; we study these effects using the FitzHugh--Nagumo model---a bistable reaction--diffusion prototype with anisotropic, spatially varying diffusion on rectangles $\Omega_\alpha$ and no-flux Neumann boundary conditions:
\begin{equation}
    \left\{
    \begin{aligned}
        \partial_t u &= \nabla \cdot (\phi_{\alpha, \beta}( \mathbf{x})  \Sigma \nabla u) + f(u)\,, \quad &\mathbf{x} \in \Omega_\alpha,\; t > 0 \\
        0 &= \left( \phi_{\alpha, \beta}( \mathbf{x}) \Sigma \nabla u \right) \cdot \mathbf{n}(\mathbf{x})\,, \quad &\forall \, \mathbf{x} \in \partial\Omega_\alpha\,, 
    \end{aligned}
    \right.
\label{FN_eq}
\end{equation}
Here $f(u)$ is cubic and governs bistable kinetics; $\Sigma$ is a diagonal tensor with fixed $2{:}1$ $x{:}y$ anisotropy that is pointwise scaled by a Gaussian well $\phi_{\alpha, \beta}(\mathbf{x})$ (random rotations and deviations, minimum at the origin). 
In this example, $\alpha$ indexes the rectangular domain, while $\beta$ indexes the field $\phi_{\alpha, \beta}$ associated with each domain, which multiplies pointwise the fixed diffusion tensor $\Sigma$. In formulas: $\mathbf{K}_{\alpha, \beta}(\mathbf{x}) = \phi_{\alpha, \beta}(\mathbf{x}) \, \Sigma$.

\paragraph{Data preparation.}
For each instance, $\Omega_\alpha$ is an axis-aligned rectangle centered at the origin with side lengths $L_x,L_y\sim\mathrm{Unif}([5,10])$, discretized on a uniform grid with spacing $0.05$; a randomly oriented Gaussian well $\phi_{\alpha, \beta}$ modulates the $2{:}1$ diagonal diffusion tensor, and a circular seed triggers bistable front propagation (SI Appendix, section~S1.3).
In total, \textbf{$510$} such domains are simulated with multiple independent trajectories each; the experiments in this paper use $500$ training trajectories (one per domain on a $500$-domain subset) and $200$ test trajectories ($20$ on each of $10$ held-out domains).
Fields are Poisson-subsampled in space (minimum radius $0.15$) with linear interpolation to the collocation points, which resulted in discretized domains of $1644.39 \pm 427.12$ points (minimum 813, maximum 2805).

See SI Appendix, section~S1.3, for further details, including parameter sampling and time integration.

\begin{figure}[!htbp]
    \centering

    \begin{subfigure}{0.9\textwidth}
        \centering
        \includegraphics[trim={10 5 8 5},clip,width=\textwidth]{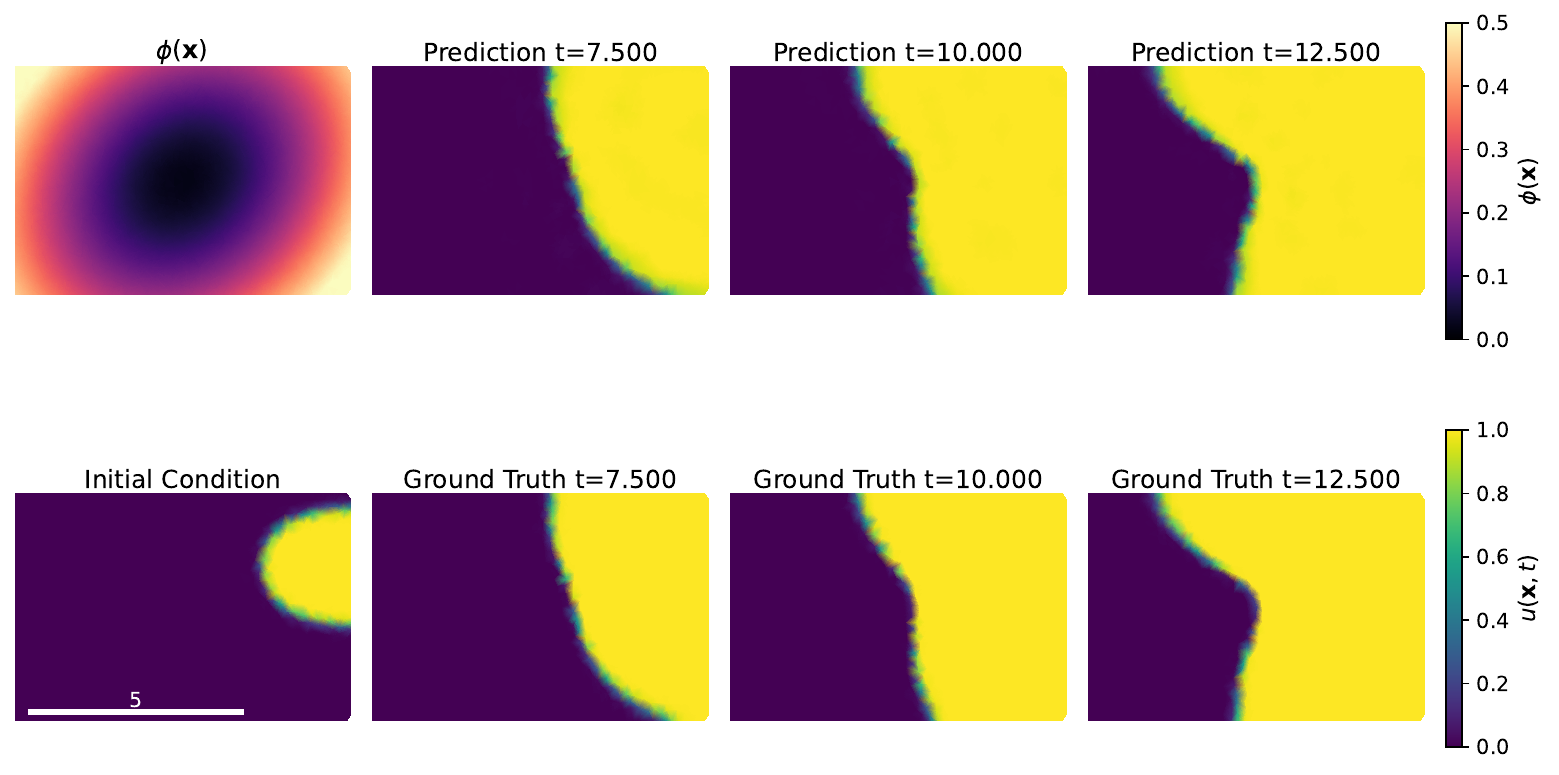}
        \caption{}
        \label{fig:fnA}
    \end{subfigure}

    \vspace{0.5em}
    \hrule
    \vspace{0.5em}

    \begin{subfigure}{0.9\textwidth}
        \centering
        \includegraphics[trim={10 5 8 5},clip,width=\textwidth]{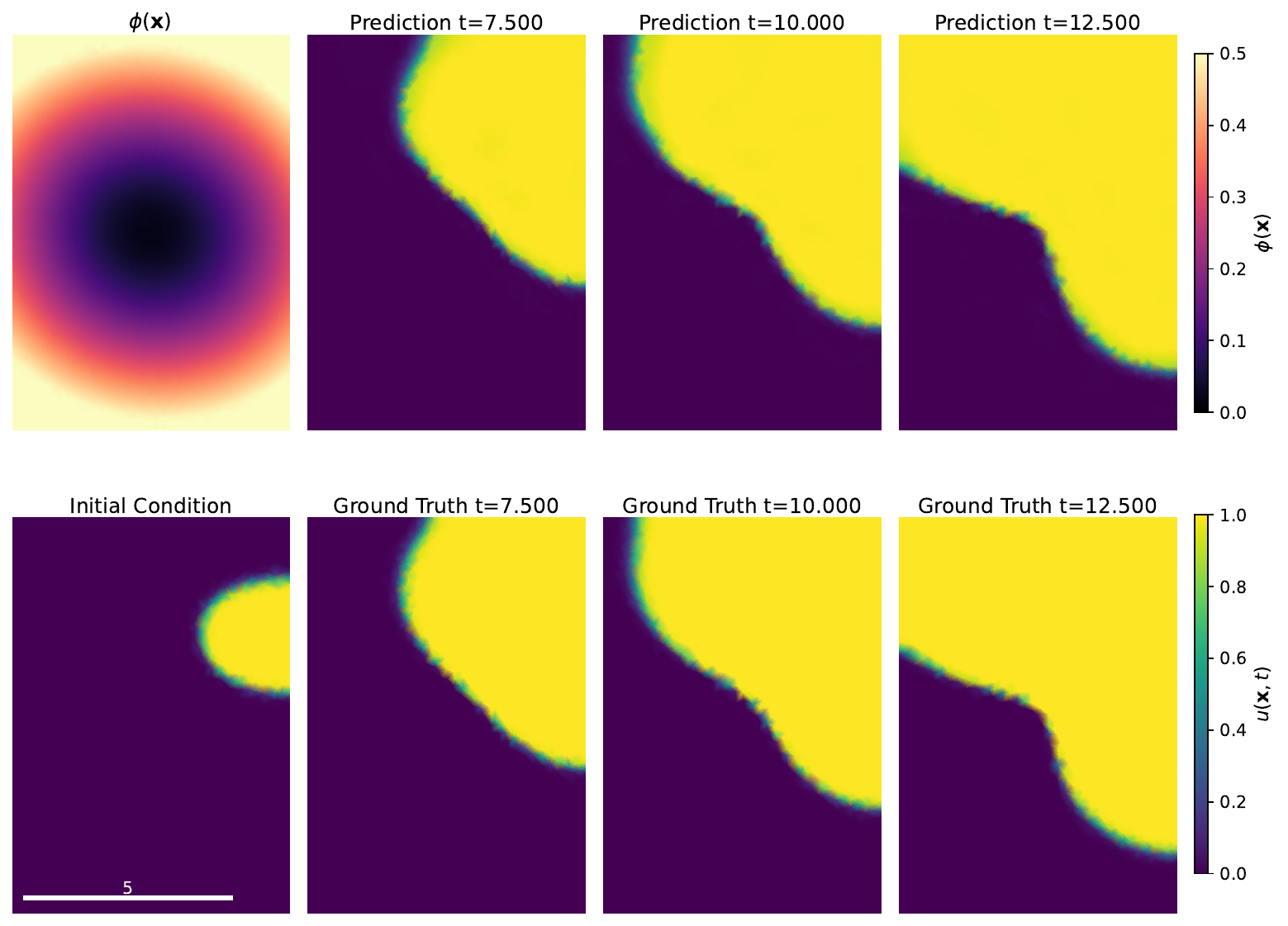}
        \caption{}
        \label{fig:fnB}
    \end{subfigure}

    \caption{\textbf{FitzHugh--Nagumo with Various Domain and Diffusive Parameters: G-FuNK vs Target (Two Test Examples).}
        Two representative test examples corresponding to different domains with different Gaussian-defined anisotropic diffusive fields. In each example, the \textit{top-left} panel shows the pointwise multiplying function $\phi_{\alpha, \beta}(\mathbf{x})$ that combines with $\Sigma$ and defines the local strength of the anisotropic (ratio 2:1) diffusive field $\mathbf{K}_{\alpha, \beta}(\mathbf{x})$, as defined in (\ref{FN_eq}). The \textit{bottom-left} panel shows the initial condition at $t_0=2.5$. The remaining panels show snapshots of the solution at three later time points, with the G-FuNK predictions shown above the corresponding ground truth targets.
        }
    \label{FN_figure}
\end{figure}

\paragraph{G-FuNK performance.}
\noindent
Figure~\ref{FN_figure} shows two representative test trajectories where G-FuNK tracks the heterogeneous diffusive fronts.
SI Appendix, Fig.~S2 illustrates Geo-FNO sensitivity to adapting to domains of varying sizes, diffusivities and initial conditions. In this example the time-averaged rel. $L^2$ error was large ($0.344$) compared to other examples. However, the standard deviation across the test set was large ($0.164$). Geo-FNO is able to accurately reproduce the solution for some examples but fails in others.

\subsubsection{Cardiac electrophysiology}

\paragraph{PDE problem.}
We target left-atrial (LA) conduction modeled as reaction--diffusion on patient-specific surfaces coupled to high-dimensional membrane kinetics \cite{Roney2021,Ali2021,Roney2019}.
On each anatomical mesh, finite-element simulations (\texttt{openCARP}; Courtemanche ionic model \cite{plank2021opencarp,courtemanche1998ionic}) are used to solve the equations, in the same form as those in Eq.~\eqref{EP_equation}, with homogeneous Neumann flux at the boundary to ensure the domain is electrically insulated from the surrounding tissue and fluids.
Conductivity follows myocardial fibers with ratio $5{:}1$ along versus across the fiber and baseline $0.625\,\text{mS}/\text{mm}$. 
In contrast to the previous example, where the conductivity tensor $\Sigma$ was fixed across all domains, each cardiac geometry is associated with a single geometry-specific diffusion tensor field $\mathbf{K}_{\alpha, 0} (\mathbf{x})$ defined by myocardial fiber orientations.
A random localized delta stimulus initiates propagation, and surrogate training starts from snapshots $10\,\text{ms}$ afterward as specified in SI Appendix, section~S1.4.
The supervised target throughout is the full spatiotemporal transmembrane potential $u(\mathbf{x},t)$, approximating the entire trajectory from the initial condition $u_0$ under this physical model.

\paragraph{Data preparation.}
Geometries are \textbf{$25$} patient-specific LA surface meshes derived from clinical CT with atlas-based myocardial fiber fields \cite{Ali2021, Roney2021}; each mesh is forward-simulated in \texttt{openCARP} and downsampled to $\approx 5000$ vertices with $\SI{5}{\milli\second}$ snapshots on $[0,\SI{85}{\milli\second}]$.
Every anatomy contributes $20$ high-fidelity trajectories ($500$ total); we use $20$ trajectories per mesh when forming train/test datasets, and we perform five single-geometry hold-out evaluations that train on \textbf{$24$} meshes per split with one geometry reserved for testing across five random splits.
See SI Appendix, section~S1.4, for tensor assembly, downsampling, and interpolation.

\begin{figure}[!htbp]
    \centering

    \includegraphics[trim={10 5 10 5},clip,width=\textwidth]{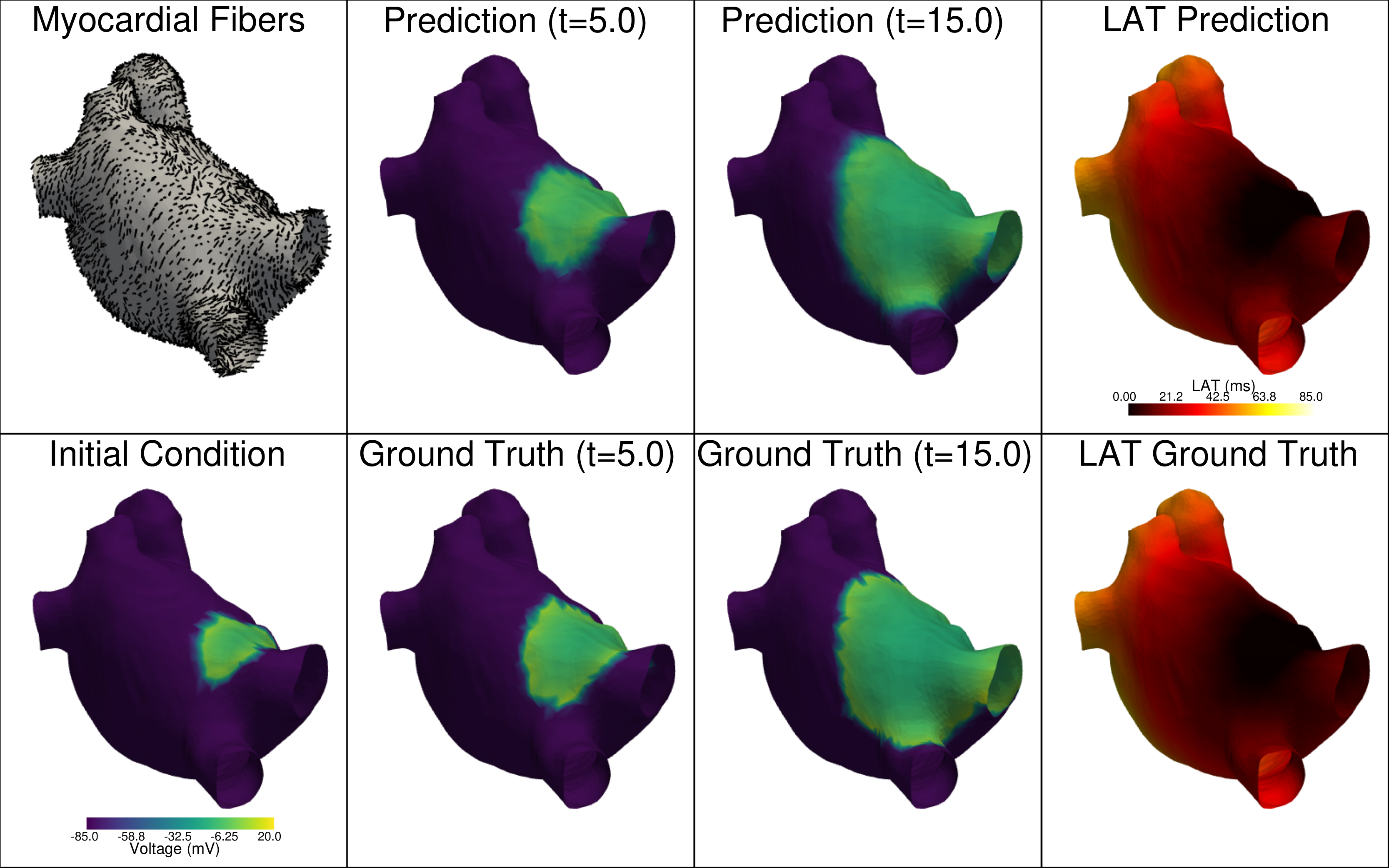}

    \caption{\textbf{Cardiac Electrophysiology: G-FuNK vs Target (Out-of-Training Test Geometry).}
    Comparison of G-FuNK predictions and ground truth targets on a representative out-of-training patient-specific atrial geometry. The \textit{top-left} panel shows the myocardial fiber field, where the black lines indicate the primary direction of electrical conduction throughout the left atrium and define the orientation of the anisotropic diffusion tensor $\mathbf{K}_{\alpha,0}(\mathbf{x)}$. The \textit{bottom-left} panel shows the initial condition of transmembrane potential for $u(\mathbf{x},t=\SI{0}{\milli\second})=u_0$. The middle columns show transmembrane potential snapshots at $t=\SI{5}{\milli\second}$ and $t=\SI{15}{\milli\second}$, with G-FuNK predictions displayed above the corresponding ground truth targets. The rightmost column shows the local activation time (LAT) functional computed from the predicted solution and from the corresponding ground truth. A full temporal roll-out example is provided in SI Appendix, Fig.~S5.}
    \label{cardiac_ep_figure}
\end{figure}

\paragraph{G-FuNK performance.}
Beyond pointwise waveform accuracy, we summarize activation via local activation times $\mathrm{LAT}(\mathbf{x})=\inf\{t:\,u(\mathbf{x},t)\ge u_{\mathrm{th}}\}$ with $u_{\mathrm{th}}=\SI{-20}{\milli\volt}$, enabling comparison with latent-operator studies that regress LAT directly \cite{yin2024dimon, Naghavi2026}.
LAT discrepancies use only nodes activating inside the rollout window---see Table~\ref{table:results} for relative $L^2$, Pearson correlation, and LAT-focused metrics aggregated over five single-geometry hold-out splits.
Training targets the interval $0$--$85\,$ms capturing steep frontal gradients; qualitative agreement on a withheld patient is depicted in Fig.~\ref{cardiac_ep_figure}. It should be noted that the network learns on very coarse spatial and temporal discretizations compared with the finite-element solver that generated the training and test data. The neural ODE integrator used a time step of $\SI{200}{\micro\second}$ which is 40 times larger than the finite-element solver used to generate the trajectories ($\SI{5}{\micro\second}$). Additionally, the spatial representations of the domains were reduced from meshes containing $>200\text{k}$ nodes to $\approx 5\text{k}$ nodes.
In terms of computational cost, G-FuNK predictions required approximately $0.346 \pm 0.0006$ seconds on an NVIDIA A4500 GPU, compared to the finite-element solver, which required on average $4.99 \pm 1.33$ minutes (min–max: 2.95–13.42 minutes) on 6 CPU cores, for an 85-ms atrial trajectory.

\subsection{Overall Results}
Altogether, G-FuNK was adaptable to all the examples discussed above with low relative error.
The results for the various tasks are listed in Table~\ref{table:results}.
Across the benchmark problems, G-FuNK achieved accuracy comparable to or better than Geo-FNO \cite{li2023fourier} and a message-passing GNN \cite{iakovlev2020learning}.
The reported metrics include the relative $L^2$ error, Pearson correlation, and LAT relative $L^2$ error, defined respectively as

\begin{align}
    \text{Relative} \, L^2 &= \frac{1}{M} \sum_{m=1}^M \frac{1}{L}\sum_{\ell=1}^L  \frac{\left\Vert u^{(m)}(\cdot, t_\ell) - \hat{u}^{(m)}(\cdot, t_\ell)\right\Vert_2}{\left\Vert u^{(m)}(\cdot, t_\ell)\right\Vert_2 +\epsilon}\,, \label{eq:time_average_rel_l2} \\
    \text{Pearson correlation}  &= \frac{1}{M} \sum_{m=1}^M  \frac{1}{n_{\alpha^{(m)}}} \sum_{i=1}^{n_{\alpha^{(m)}}}  \rho\left(u^{(m)}(x_i, \cdot) , \hat{u}^{(m)}(x_i, \cdot) \right)\,, \\
    \text{LAT Relative} \, L^2 &= \frac{1}{M} \sum_{m=1}^M \frac{\left\Vert \mathrm{LAT}^{(m)} - \widehat{\mathrm{LAT}}^{(m)}\right\Vert_2}{\left\Vert \mathrm{LAT}^{(m)}\right\Vert_2 +\epsilon}\,.  
\end{align}

Here, \(\rho\) denotes the Pearson correlation coefficient computed between the predicted and target temporal signals at each spatial location and subsequently averaged across spatial locations. This metric measures similarity in the temporal waveform and is invariant to additive offsets and positive rescaling of signal amplitude, but remains sensitive to differences in temporal alignment.

Time-dependent relative $L^2$ errors are shown in SI Appendix, Fig.~S3.

\begin{table*}[t]
\centering
\small
\setlength{\tabcolsep}{4pt}

\begin{tabularx}{\linewidth}{@{}Xccc@{}}
\toprule
\textbf{Problem} & \textbf{G-FuNK} & \textbf{Geo-FNO} & \textbf{GNN} \\
\midrule
\textbf{Anisotropic heat} & & & \\
Parameter adaptation & $0.056 \pm 0.018$ & $0.064 \pm 0.025$ & $0.298 \pm 0.075$ \\[3pt]
\textbf{Reaction--diffusion} & & & \\
Domain adaptation & $0.100 \pm 0.025$ & $0.112 \pm 0.015$ & $0.335 \pm 0.010$ \\[3pt]
\textbf{FitzHugh--Nagumo} & & & \\
Domain + parameter adaptation & $0.046 \pm 0.017$ & $0.344 \pm 0.164$ & $0.448 \pm 0.139$ \\
\bottomrule
\end{tabularx}
\par\medskip
\begin{tabularx}{\linewidth}{@{}XrXr@{}}
\toprule
\multicolumn{4}{c}{\textbf{Cardiac electrophysiology: G-FuNK only}} \\
\multicolumn{4}{c}{Five single-geometry hold-out splits; domain + parameter adaptation} \\
\midrule
State rel.\ $L^2$ & $0.160 \pm 0.013$ & LAT MAE & $1.732 \pm 0.221~\mathrm{ms}$ \\
Pearson correlation & $0.949 \pm 0.016$ & LAT rel.\ $L^2$ & $0.045 \pm 0.004$ \\
\bottomrule
\end{tabularx}

\caption{\textbf{Prediction performance across benchmark problems.}
Performance of G-FuNK, Geo-FNO, and a GNN equipped with a neural ODE on the
problems in Section~\ref{sec2}. Reported values are relative $L^2$ errors on
held-out test sets with unseen initial conditions, domains, and, where applicable,
diffusion fields, unless otherwise indicated; where $\pm$ appears for the first
three benchmarks, the spread is over test trajectories. Cardiac electrophysiology
used five single-geometry hold-out splits; entries are mean $\pm$ standard
deviation across splits, with Pearson correlation, LAT relative $L^2$, and mean
absolute error (MAE, ms) reported in addition to relative $L^2$ error.}
\label{table:results}
\end{table*}

\FloatBarrier

\section{Discussion}\label{sec12}

\paragraph{Significance.} G-FuNK predicts the full spatiotemporal trajectory of a stiff, fiber-anisotropic reaction--diffusion system across a family of distinct patient-specific surfaces, each with its own anisotropic conduction tensor, at a cost linear in mesh size and without a coordinate map between domains. To our knowledge, earlier neural PDE surrogates have not done all of these together. Spectral graph operators are not new \cite{defferrard2016convolutional,bo2023survey,wang2022powerful,yang2022new}. G-FuNK differs in how geometry and anisotropy enter the model: the spectral basis comes from a Laplacian whose weights are set by the local diffusion tensor, so the eigenbasis carries both the shape of the domain and the direction of transport. The learned weights therefore act on a representation that already contains this information.

The graph spectral representation depends on the operator, not on the coordinate system in which the domain is embedded. We trained on one family of problems and tested on an equivalent dataset obtained by rotating the entire problem by 90 degrees. Geo-FNO, whose spectral representation depends on a global coordinate parameterization, rose to a relative $L^2$ error of 0.450 on the rotated set (SI Appendix, Fig.~S4), while G-FuNK returned the same predictions as before, because its basis is defined on the graph rather than on an embedding grid. 
The same property lets G-FuNK recover the effect of strong anisotropy from the Laplacian eigenpairs alone. Geo-FNO, by contrast, was provided the pointwise diffusion information explicitly but remained sensitive to the global coordinate representation under rotation.

G-FuNK operates directly on the intrinsic graph representation, eliminating the need for either a diffeomorphic mapping to a canonical domain or an embedding into a surrounding Cartesian grid. The human left atrium, with its pulmonary vein and mitral valve orifices, cannot be mapped smoothly to a cube, torus or sphere. Coordinate-based approaches reach it through parameterizations such as Universal Atrial Coordinates, which separate points that are adjacent on the anatomy, and Zhou et al. \cite{zhou2026unifiedframeworkgeometryindependentoperator} report elevated errors along the resulting seam. In this study, Universal Atrial Coordinates were used upstream to assign fiber orientations, but the learned operator does not use them. G-FuNK works directly on the surface graph, and on held-out patients, the low activation-time error of 1.73 ms and high Pearson correlation of 0.949 indicate close agreement in activation timing and temporal waveform despite the larger full-state relative \(L^2\) error.

A recent atrial operator-learning study allows a further, indirect comparison. Zhou et al. \cite{zhou2026unifiedframeworkgeometryindependentoperator} predict local activation-time maps across diverse patient-specific left atrial geometries using a Universal Atrial Coordinates representation, whereas G-FuNK predicts the full spatiotemporal transmembrane-potential trajectory and derives activation time from the predicted dynamics. G-FuNK achieves a LAT mean absolute error of $1.732\pm0.221~\mathrm{ms}$, compared with the $5.1~\mathrm{ms}$ MAE reported by Zhou et al. \cite{zhou2026unifiedframeworkgeometryindependentoperator}. These results were obtained on different datasets and using different evaluation protocols and therefore do not constitute a direct quantitative comparison. Nevertheless, the results show that deriving activation time from full-field spatiotemporal prediction can achieve LAT errors within the range, and in this case below the magnitude, reported by approaches designed specifically for LAT prediction, while retaining access to the complete electrophysiological trajectory.

These points explain the pattern across benchmarks in Table 1. On regular domains, where both the graph- and Fourier-based spectral representations are well suited to the underlying operator, G-FuNK and Geo-FNO perform similarly on the anisotropic heat and rectangular reaction-diffusion problems, with Fourier spectral methods appearing as a special case of the broader graph spectral representation. G-FuNK is more accurate where the grid and coordinate assumptions do not hold well: under spatially varying, non-periodic anisotropy (FitzHugh-Nagumo, 0.046 against 0.344), under rotation of the entire PDE, and on the atrial surface, where the Geo-FNO formulation considered here is not directly applicable. Message-passing GNNs have demonstrated strong performance for PDE learning in prior work. In the present benchmarks, however, the evaluated message-passing GNN produced substantially larger errors than G-FuNK and Geo-FNO across all three examples, indicating that the considered local message-passing architecture was less effective under these discretization and training settings. To keep the comparison fair, we ran Geo-FNO with the same neural ODE integrator used for G-FuNK, since its published implementation does not produce full trajectories on its own.

Electrophysiology simulations are known to be sensitive to the spatial and temporal resolution of the solver \cite{Niederer2011}. G-FuNK reproduces the dominant dynamics on much coarser graphs and at much larger time steps, up to 40 times the step of the finite-element solver in our experiments, while keeping its error low. The learned operator therefore remained accurate over the spatial and temporal resolutions used here.

\paragraph{Clinical implications.} Digital twins are becoming part of precision cardiology \cite{Niederer2019, trayanova2024computational}, but patient-specific electrophysiology simulations are expensive, and building a personalized model still requires calibration against sparse and uncertain clinical measurements \cite{Zappon2026}. In this setting, the main value of G-FuNK is speed: it can predict spatiotemporal electrophysiology dynamics quickly and repeat predictions across many geometries, parameter fields, and initial conditions. This capability could support hypothesis generation, mechanistic analysis of activation dynamics, and large parameter sweeps in patient-specific models. In particular, the ability to rapidly evaluate activation across different anatomies provides a potential computational tool for exploring how patient-specific structure and electrophysiological properties influence wave propagation. G-FuNK is not intended to replace high-fidelity biophysical solvers, but demonstrates that graph spectral operators can serve as fast surrogates for patient-specific electrophysiology.

\paragraph{Limitations.} Graph construction is an important design choice. Empirically, we observed that increasing the connectivity radius improved performance in our experiments, producing denser and smoother Laplacians that favored lower-frequency modes. However, this bias may limit the model's ability to capture fine-scale or high-frequency features in strongly heterogeneous diffusion fields, and a fuller study of this trade-off is left for future work. The choice of graph construction, such as the connectivity radius or the nearest-neighbor rule, is itself a design decision that can affect performance from one domain to another. The choice of Laplacian normalization also influences performance. While the random walk Laplacian provides a natural density correction and is commonly used in diffusion map formulations, the symmetric normalized Laplacian yielded improved performance in our experiments. This may be related to the orthogonality properties of its eigenvectors and the resulting stability of the spectral representation. An additional comparison between Laplacian normalizations and edge weighting schemes is provided in SI Appendix, section~S4. An additional concern is the variability of the spectral basis across domains. When eigenvalues lie close together, the ordering of the corresponding eigenmodes can change, which can add to the prediction error. Aligning or matching eigenvectors across domains could address this.

\paragraph{Conclusion.} For tasks such as patient-specific parameter sweeps and mechanistic studies, reducing simulation cost from minutes per trajectory to under a second could enable substantially larger computational studies across geometries, parameter fields, and initial conditions. Beyond speed, G-FuNK is defined directly on graphs, allowing it to operate on complex geometries without structured grids or global parameterizations. Across the benchmarks, G-FuNK performs comparably to Geo-FNO on problems where Fourier representations are effective and provides greater accuracy when anisotropy and geometric complexity challenge coordinate-based representations. In the cardiac electrophysiology example, G-FuNK predicts the spatiotemporal transmembrane potential across complex patient-specific anatomies, enabling local activation times to be derived with accuracy comparable to previously reported neural operator approaches. These results demonstrate that graph spectral operator learning provides a general and practical approach for learning PDE solution operators on irregular domains.

\FloatBarrier
\newpage

\section{Methods}\label{sec11}

The operator-learning formulation (the flow map induced by Eq.~\ref{general_diff_op}, the solution operator, and the learned one-step update and flow) is given in SI Appendix, section~S9.

\noindent \textbf{Reduced Models and Homogenization}.
In fact, we are very much interested in situations where the ``true'' system is driven by equations more general than \eqref{general_diff_op}, with many more variables, possibly evolving at multiple time scales; this will be the case in examples 2 and 4, where besides the spatial variables and the unknown action potential $u$, there are many other quantities $\mathbf{v}$ evolving in space and time (e.g. modeling ionic concentrations that affect the evolution of $u$, and these are driven by a large system of ODEs). 
In these situations, given observations from such systems, our estimated equations (SI Appendix, Eq.~S15) can be viewed as performing dimension reduction on the variables (e.g. by keeping only the spatial variables) and homogenization, obtaining an effective equation for $u$, without tracking the evolution of the other quantities $\mathbf{v}$. Accordingly, for these examples \(\mathcal{N}_\theta\) should be interpreted as an effective closure for the evolution of \(u\) over the distribution of trajectories represented in the training data, rather than as a globally valid autonomous generator of the full ionic system.

\subsection[G-FuNK: Graph Fourier Neural Kernel]{G-FuNK: \underline{G}raph \underline{F}o\underline{u}rier \underline{N}eural \underline{K}ernel}
\subsubsection{Graphs, Laplacians, and the Graph Fourier Transform}\label{graph_laplacian_stuff}
Let $\mathcal{G}(\mathcal{V},\mathcal{E})$ be an undirected graph where $\mathcal{V}=\{x_i\}_{i=1}^{n_{\alpha}}$ is the set of \(n_{\alpha}\) nodes (or vertices) constructed from a finite-element mesh discretization or downsampled point cloud representation of $\Omega_{\alpha}$ and \( \mathcal{E} \subseteq \{ (x_i, x_j) \mid x_i, x_j \in \mathcal{V} \} \) is the set of edges, where a pair of nodes $(x_i,x_j)$ is connected by an edge if $x_j$ is a $k$-nearest neighbor or within a certain distance of each $x_i$.
Let \( W\in\mathbb{R}^{n_{\alpha}\times n_{\alpha}} \) be the weighted adjacency matrix, with $W_{ij}=W_{ji}\ge0$, with strict inequality if and only if $(x_i,x_j)\in\mathcal{E}$. 
\( D \) is the $n_{\alpha}\times n_{\alpha}$ diagonal degree matrix of the graph, defined by $D_{ii}:=\sum_j W_{ij}$, and lastly, a symmetric normalized graph Laplacian is constructed which is a positive semi-definite matrix with a spectral decomposition given by 
\begin{align}
\mathscr{L} := \mathrm{I} - D^{-1/2} W D^{-1 / 2} = \Psi \Lambda \Psi^T,
\end{align}
where \( \Psi = [\psi_1, \ldots, \psi_{n_\alpha}]\in\mathbb{R}^{{n_{\alpha}}\times {n_{\alpha}} }\) is the matrix of orthonormal eigenvectors of \(\mathscr{L}\), and \( \Lambda = \operatorname{diag}(\lambda_1, \lambda_2, \ldots, \lambda_{n_{\alpha}}) \) is the diagonal matrix of the corresponding non-negative eigenvalues.
Note that $\{\psi_i\}_{i=1}^{n_{\alpha}}$ is an orthonormal basis for functions on the graph, since $\mathscr{L}$ is symmetric, and the eigenvalue $0$ has multiplicity equal to the number of connected components of the graph, which we can assume to be 1 without loss of generality. $\mathscr{L}$ is a sparse matrix, and when we only compute the $k_{\max}$ smallest eigenvalues and eigenvectors, we can use a sparse symmetric iterative eigensolver.

When the graph is constructed on points randomly sampled on a domain or a manifold, this discrete Laplacian is an approximation (in a suitable sense) to a continuous Laplacian in the limit as the number of samples goes to infinity, and its eigenvectors are approximations to the eigenfunctions of the continuous Laplacian, which are the natural generalization of Fourier modes from a torus or rectangular box to general domains and manifolds. Furthermore, the geometry of the domain (graph, in the discrete case, Riemannian manifold in the continuous case) is completely determined by the Laplacian, and therefore by its eigenvalues and eigenvectors. 

We define the Graph Fourier Transform (GFT) as follows: for a function $a:\mathcal{V}\rightarrow\mathbb{R}^{d_{u}}$ defined on the nodes of \( \mathcal{G} \), the Graph Fourier Transform (GFT), denoted by \( \mathcal{F} \), rewrites \( a \) in the basis of eigenvectors of the graph Laplacian \( \mathscr{L} \), yielding a function $\hat{a}:\Lambda\rightarrow\mathbb{R}^{d_{u}}$ of the eigenvalues $\lambda_1,\dots,\lambda_{n_{\alpha}}$, defined as
\begin{equation}
\hat{a} := \mathcal{F}a := (\langle \psi_i, a \rangle)_{i=1,\dots,{n_{\alpha}}}=\Psi^Ta\,,
\end{equation}
with the inner product computed on each component of the vector-valued function $a$.
In the reverse process, the inverse Graph Fourier Transform (IGFT), represented by \( \mathcal{F}^{-1} \), reconstructs the function \( a \) from its spectral representation \( \hat{a} \), through the equation
\begin{equation}
a = \mathcal{F}^{-1}\hat{a} := \sum_{i=1}^{n_{\alpha}} \hat a(\lambda_i)\psi_i = \Psi \hat{a}\,.
\end{equation}
It should be noted that computing the $k_{\max}$ lowest eigenvalues of $\mathscr{L}$ is inexpensive, with a complexity of $O(n_\alpha k_{\max}^2)$ for a graph with $n_{\alpha}$ vertices using sparse symmetric matrix eigensolvers. Of course, this calculation is performed per domain/diffusivity parameter, rather than being recomputed at every instance during network training.

Graph diffusion is analogous to the heat equation and describes the diffusive movement of a function across a graph. 
The solution to the graph diffusion equation can be expressed in the basis of the spectral decomposition of the graph Laplacian $\mathscr{L}$ as
\begin{equation}
    \frac{\partial a(t)}{\partial t} = - c \,\mathscr{L} \, a(t) \quad \rightarrow \quad 
    a(t) = \sum_{i=1}^{n_{\alpha}} \mathcal{F}(a(0))_i e^{-c \lambda_i t} \psi_i
    \label{graphdiffeq_spectralgraphlaplaciansolution}
\end{equation}
where $c$ is the diffusivity constant, and $a(t)\in\mathbb{R}^{n_{\alpha}}$ is the function at time $t$, and the initial condition is the function $a(0)=(a_i(0))_{i=1}^{n_{\alpha}}$. The solution can be truncated to the $k_{\max}$ lowest modes, where $k_{\max} \leq n_{\alpha}$, for an approximated solution. It should be noted that the {\it{form}} of Eq. (\ref{graphdiffeq_spectralgraphlaplaciansolution}) is not dependent upon a spatial coordinate but only on the spectral decomposition of $\mathscr{L}$.

By establishing these preliminaries, we set the stage for the development of G-FuNK, which leverages the spectral properties (precomputed lowest eigenpairs) of the graph Laplacian to efficiently learn and generalize dynamic processes across different domains.
\subsubsection{The G-FuNK layers and network}
In this section, we describe the neural operator framework, a variation of the FNO of \cite{li2020fourier}, equipped with our novel \textbf{G}raph \textbf{F}o\textbf{u}rier \textbf{N}eural \textbf{K}ernel (G-FuNK) layers to learn solution generators to time-dependent PDEs across multiple domains and multiple anisotropic diffusion tensor fields.

\begin{figure}[!htbp]
    \centering
\includegraphics[trim={10 100 8 100},clip,width=.8\textwidth]{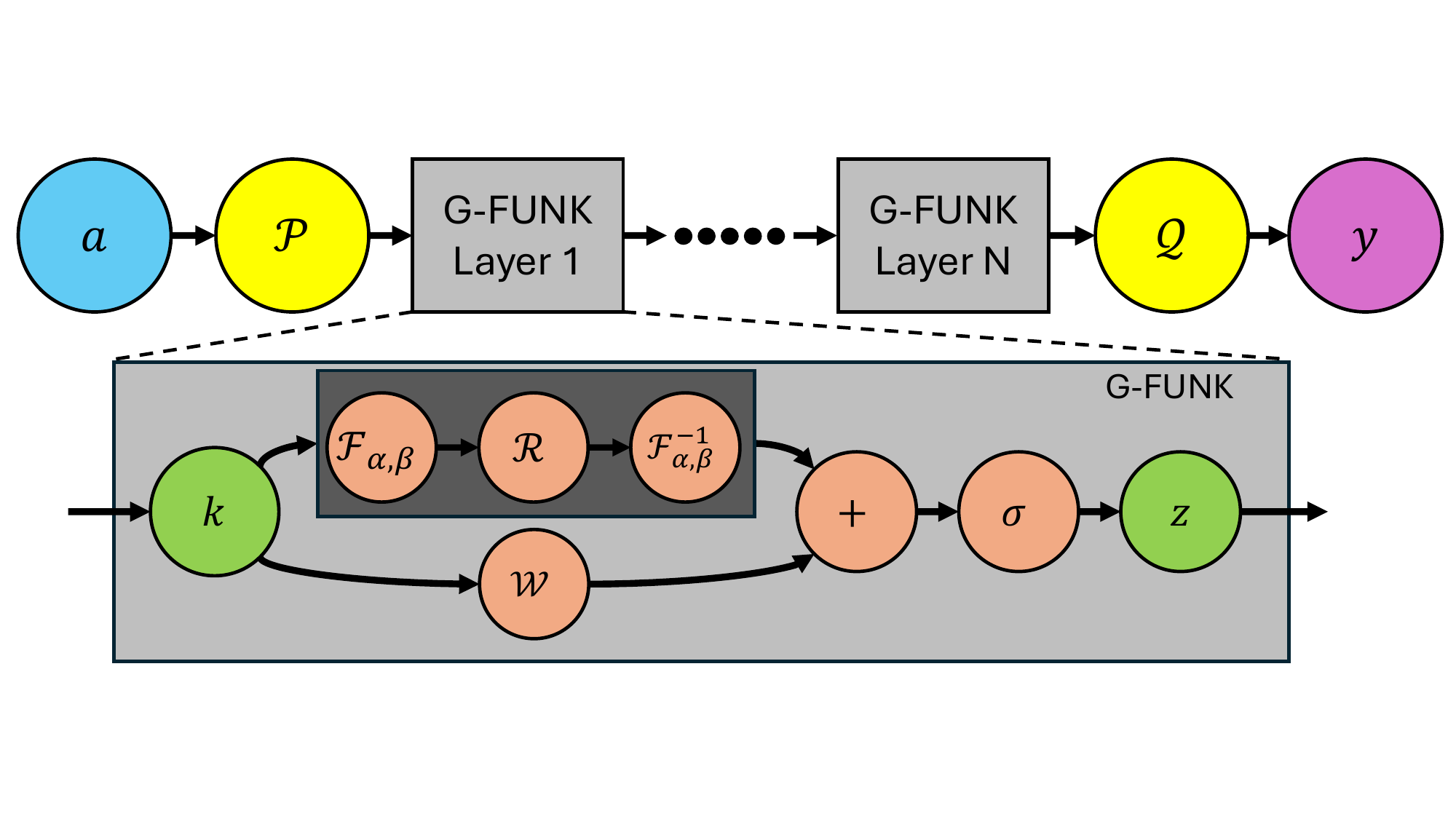}
    \caption{\textbf{Top:} Network structure of the Neural Operator. The input $a$ is a (possibly vector-valued) function on $\mathcal{G}_{\alpha,\beta}$. (1) It is lifted to a higher dimensional space via a lifting operator $\mathcal{P}$. (2) Multiple G-FuNK layers are then applied. (3) The network projects to the output dimensional space via $\mathcal{Q}$.
\textbf{Bottom:} Single G-FuNK layer structure. Starting from input $k$, \textbf{Spectral branch} (1) $\mathcal{F}$ denotes the GFT of the input which is reduced to the $k_{\max}$-lowest modes. 
(2) The function is transformed with a learned spectral operator $\mathcal{R}$ which incorporates the eigenvalues.
(3) $\mathcal{F}^{-1}$ represents the inverse GFT, mapping back to the original domain. 
(4) The input function $k$ undergoes a linear mapping in $\mathcal{W}$. 
(5) The outputs of the top and bottom branches are combined and then passed through an activation function $\sigma$. The output of the G-FuNK layer is $z$ and is operated on by the next G-FuNK layer.}
\label{gfunkdiagram}
\end{figure}

Recall that the domain $\Omega_\alpha$ -- or, rather, a discretization thereof -- is given, and so is the diffusion field $\mathbf{K}_{\alpha,\beta}$.
We exploit the knowledge of the former to construct a graph $\mathcal{G}_{\alpha, \beta}$ which discretizes $\Omega_{\alpha}$ into $n_{\alpha}$ nodes $\mathcal{V}=\{x_i\}_{i=1}^{n_{\alpha}}$, and knowledge of the latter to weight the edges in order to approximate the second-order term $\nabla\cdot(\mathbf{K}_{\alpha, \beta}(\mathbf{x})\nabla)$ in the PDE. This is where we use the fact that the PDE is semilinear. By choosing, on each edge $\mathcal{E}_{ij}$,

\begin{equation}
W_{ij}
=
\frac{1}
{D_{\mathrm{weight}}}
\exp\!\left(
-
\frac{
(x_j-x_i)^T
\mathbf{K}_{ij}^{-1}
(x_j-x_i)
}{
2\sigma^2
}
\right),
\label{heat_kernel_eq}
\end{equation}
we weight the edges using an anisotropic Gaussian kernel, where the Mahalanobis distance is defined by the locally averaged inverse diffusion tensor 
$\mathbf{K}_{ij}^{-1} = \frac12 \left( \mathbf{K}_{\alpha, \beta}^{-1} (x_i) + \mathbf{K}_{\alpha, \beta}^{-1} (x_j) \right)$, 
with $D_{\mathrm{weight}} = \left(\det(\mathbf{K}_{\alpha,\beta}(x_i))\det(\mathbf{K}_{\alpha,\beta}(x_j))\right)^{1/4}$ 
and the bandwidth was chosen as $\sigma = \frac r2$, where $r$ is the graph connection radius. Consequently, neighboring nodes aligned with directions of high diffusivity receive larger edge weights than equally distant nodes in low-diffusivity directions. For non-planar geometries (cardiac EP), the Euclidean displacement is replaced by the geodesic distance $l_{ij}$ and the local geodesic tangent direction $\hat{r}_{ij}$, yielding 
$W_{ij}=\frac{1}{D_{\mathrm{weight}}}\exp\!\left(-\frac{l_{ij}^2 \hat{r}_{ij}^T\mathbf{K}_{ij}^{-1}\hat{r}_{ij}}{2\sigma^2}\right)$, 
where $\hat{r}_{ij}$ denotes the unit tangent vector of the shortest geodesic connecting vertices $i$ and $j$, and $l_{ij}$ is the corresponding geodesic distance.
For all $\mathcal{G}_{\alpha, \beta}$ we ensure an undirected graph by symmetrizing the edge weights between connected nodes. 

Each unique parameter set $(\Omega_{\alpha}, \mathbf{K}_{\alpha, \beta}) \in \mathcal{M}$ results in a unique weighted graph $\mathcal{G}_{\alpha, \beta}$, a corresponding Laplacian $\mathscr{L}_{\alpha,\beta}$, and a corresponding set of eigenpairs, giving $\Psi_{\alpha, \beta}$ and $\Lambda_{\alpha, \beta}$. We will be only computing and using the $k_{\max}$ lowest frequency eigenvectors, corresponding to the $k_{\max}$ smallest eigenvalues; with some abuse of notation, we denote these again by $\Psi_{\alpha,\beta}$ and $\Lambda_{\alpha,\beta}$. Providing these eigenpairs (truncated to the $k_{\max}$ lowest modes) permits the G-FuNK model to project signals on various-sized graphs into a common spectral representation and map them back to the graph domain using the GFT and IGFT, $\mathcal{F}_{\alpha, \beta}$ and $\mathcal{F}_{\alpha, \beta}^{-1}$, respectively.

The neural operator $\mathcal{N}_{\theta}$, equipped with several G-FuNK layers, approximates the solution generator of the PDE by applying a series of transformations, which can be formally expressed as the composition of different functions representing the network layers. The network begins with the following function transformation:
\begin{align*}
    k_0 := \mathcal{P}(a; \theta_{\mathcal{P}}), 
\end{align*}
where $a:\mathcal{V}\rightarrow\mathbb{R}^{d_u}$ is a vector-valued
function defined on the vertices
$\mathcal{V}=\{x_i\}_{i=1}^{n_\alpha}$ of
$\mathcal{G}_{\alpha,\beta}$, and $\mathcal{P}$, parameterized by
$\theta_{\mathcal{P}}$, maps the node-wise input $a$ to the latent
representation
$k_0:\mathcal{V}\rightarrow\mathbb{R}^{d_p}$, typically with
$d_p\gg d_u$; the $d_p$ components of $k_0$ are referred to as channels.

For each G-FuNK layer $n = 1, \ldots, N$, we have
\begin{equation}
\begin{aligned}
    \hat{k}_n(\lambda)
    &:= \mathcal{F}_{\alpha,\beta}
    (k_{n-1};\Psi_{\alpha,\beta})(\lambda), \\
    \hat{r}_n(\lambda)
    &:= \mathcal{R}_n
    (\hat{k}_n;\theta_{\mathcal{R}_n},
    \Lambda_{\alpha,\beta})(\lambda), \\
    f_n(x)
    &:= \mathcal{F}_{\alpha,\beta}^{-1}
    (\hat{r}_n;\Psi_{\alpha,\beta})(x), \\
    \omega_n(x)
    &:= \mathcal{W}_n
    (k_{n-1};\theta_{\mathcal{W}_n})(x), \\
    z_n(x)
    &:= \sigma\!\left(f_n(x)+\omega_n(x)\right), \\
    k_n(x)
    &:= z_n(x).
\label{gfunk_operations_lambda}
\end{aligned}
\end{equation}
where $\mathcal{F}_{\alpha,\beta} :
\mathbb{R}^{n_{\alpha} \times d_p}
\rightarrow
\mathbb{R}^{k_{\max} \times d_p}$
is the (linear, partial) GFT that projects and represents the function on the
$k_{\max}$ lowest-frequency graph Fourier modes.

$\mathcal{R}_n :
\mathbb{R}^{k_{\max} \times d_p}
\rightarrow
\mathbb{R}^{k_{\max} \times d'_p}$,
for $n=1,\ldots,N$, is a learned spectral operator acting mode-wise on the
spectral coefficients. In this work, $d_p=d'_p$, and $\mathcal{R}_n$ acts
diagonally in the graph Fourier basis. Its mode-wise spectral multipliers are
generated from the corresponding graph Laplacian eigenvalues by a multi-head
self-attention network parameterized by $\theta_{\mathcal{R}_n}$.

Specifically, let
\[
\boldsymbol{\lambda}_{\alpha,\beta}
=
(\lambda_1^{\alpha, \beta},\ldots,\lambda_{k_{\mathrm{max}}}^{\alpha, \beta})^\top
\]
denote the sequence of the $k_{\mathrm{max}}$ lowest graph Laplacian
eigenvalues. The mode-wise multipliers are given by
\[
m_n(\boldsymbol{\lambda}_{\alpha,\beta})
=
\operatorname{Attn}_{\theta_{\mathcal{R}_n}}
(\boldsymbol{\lambda}_{\alpha,\beta})
\in \mathbb{R}^{k_{\mathrm{max}}},
\]
and the spectral operation is
\[
\hat{r}_n
=
\operatorname{diag}
\left(m_n(\boldsymbol{\lambda}_{\alpha,\beta})\right)
\hat{k}_n .
\]

$\mathcal{F}_{\alpha,\beta}^{-1} :
\mathbb{R}^{k_{\max} \times d'_p}
\rightarrow
\mathbb{R}^{n_{\alpha} \times d'_p}$
is the (linear) IGFT that maps the transformed spectral function back to the
graph domain.
$\mathcal{W}_n :
\mathbb{R}^{n_\alpha \times d_p}
\rightarrow
\mathbb{R}^{n_\alpha \times d'_p}$,
for \(n=1,\ldots,N\), are additional linear mappings applied
pointwise to $k_{n-1}$;
$\sigma : \mathbb{R}^{n_\alpha \times d'_p} \rightarrow \mathbb{R}^{n_\alpha \times d'_p}$ is an activation function applied, entry-wise, in each G-FuNK layer.

After passing through all $N$ G-FuNK layers, the final output is obtained as
\begin{align}
    y = \mathcal{Q}(z_N; \theta_{\mathcal{Q}})\,,
\end{align}
where $\mathcal{Q} : \mathbb{R}^{n_\alpha \times d'_p} \rightarrow \mathbb{R}^{n_\alpha \times d_v}$ maps from the last latent vector-valued space to an output vector space (all examples in this work used a scalar output space). 
Altogether, the neural operator with $N$ G-FuNK layers can be expressed as the following composition of the aforementioned transformations to approximate \(\partial_t u(\mathbf{x},t)\):

\begin{equation}
\mathcal{N}_{\theta}(a) := \mathcal{Q} \circ \underbrace{\sigma(\mathcal{W}_N + \mathcal{F}_{\alpha,\beta}^{-1} \circ  \mathcal{R}_N \circ \mathcal{F}_{\alpha,\beta} )}_{\text{G-FuNK Layer $N$}} \circ \cdots \circ \underbrace{\sigma(\mathcal{W}_1 + \mathcal{F}_{\alpha,\beta}^{-1} \circ \mathcal{R}_1 \circ  \mathcal{F}_{\alpha,\beta})}_{\text{G-FuNK Layer $1$}}  \circ \mathcal{P}(a),
\end{equation}

We denote \(\theta = \{\theta_{\mathcal{P}}, \theta_{\mathcal{R}_n}, \theta_{\mathcal{W}_n}, \theta_{\mathcal{Q}}\}_{n=1}^{N}\) as the collection of all learnable parameters in the network, which are optimized during training to minimize the discrepancy between trajectories obtained by integrating the learned generator and the corresponding numerically computed reference trajectories.

In the G-FuNK framework, the eigenvectors of the Laplacian perform the same action as the Fourier transform in the FNO framework, with the important difference that these eigenvectors are adapted to the diffusion coefficient, capturing the parametric dependence of the highest order differential term in the PDE. We suggest using the true eigenfunctions of the shape, which embed global structural and physical properties, as corroborated by numerical and statistical methods that represent PDE solutions in terms of known basis functions that contain information about the solution structure \cite{bhattacharya2021model,nagy1979modal,almroth1978automatic}. 
In FNOs, the Fourier basis is fixed across domains, and the eigenvalues are therefore not explicitly used in constructing the spectral multipliers; here, incorporating the eigenvalues allows the spectral representation to account for both the dependence on $\mathbf{K}_{\alpha, \beta}$ and on the domain $\mathcal{G}_{\alpha, \beta}$. The self-attention mechanism provides a nonlinear mapping from the eigenvalues to the spectral multipliers, departing from the classical architecture of FNOs, in which the Fourier-mode multipliers are learned independently of the input spectrum. This construction is used throughout our experiments.

Finally, we comment that $\mathcal{R}_n$ allows G-FuNK to approximate spectral multipliers, while $\mathcal{W}_n$ allows for the approximation of spatial pointwise multipliers. At a high level, this is particularly well-suited for reaction-diffusion equations, where the diffusion component is expected to be easy to learn through spectral multipliers, and the nonlinear reaction terms act pointwise and can be expected to be captured by the spatial multipliers. Of course, such PDEs have complex interactions between the spatial and spectral domains, which is what the multiple layers connected by nonlinearities aim to capture.

\noindent\textbf{Mesh-resolution transferability of G-FuNK}.\par\smallskip
Although the proposed framework involves discretization-dependent steps, such as forming a weighted graph on the sampled domain, the learned operator can be evaluated on graph representations with different node counts. As the spatial sampling is refined, the graph Laplacian provides an increasingly resolved approximation of the underlying continuous operator, and established methods exist for relating graph eigenfunctions to their continuous counterparts \cite{DiffusionPNAS}. To empirically assess sensitivity to spatial discretization, we evaluate G-FuNK across multiple graph resolutions without retraining. As shown in SI Appendix, Table~S3, performance remained stable over a range of approximately 4,000--10,000 nodes surrounding the training resolution of approximately 5,000 nodes, with increased error observed at the coarsest resolution of approximately 2,500 nodes. These results demonstrate useful resolution transferability over the range tested, while also indicating that sufficiently dense spatial sampling remains important for accurately representing the graph operator.

\noindent\textbf{Integration in time and training}.
Given discrete observations $\{u(x_i,t_\ell)\}_{i=1,\ell=1}^{n_{\alpha}, \,L}$ of the state variable at a location \( x_i \in \mathcal{G}_{\alpha, \beta} \) at different time steps $0=t_1<\dots<t_L$, we can integrate the approximation over time using an ODE solver, $\mathcal{I}$:
\begin{align}
    u(x_i,t_{\ell}+\Delta t) = \mathcal{I}\left(\mathcal{N}_{\theta}(u(x_i,t_{\ell}), \Psi_{\alpha, \beta}, \Lambda_{\alpha, \beta})\right), 
    \label{time_integration} 
\end{align}
where \( \Delta t \) is the time step size. In this work we use a neural ODE numerical algorithm (in particular, the one proposed by \cite{chen2018neural}). 

Throughout this work, since the objective is to learn the full spatiotemporal trajectory, the loss function is defined as the mean squared error between predicted and ground-truth observations, averaged over sampled spatial locations and discrete time points:
\begin{equation}
    J(\theta)=\frac{1}{M} \sum_{m=1}^M \frac{1}{L}\sum_{\ell=1}^L \frac{1}{n_{\alpha^{(m)}}} \sum_{i=1}^{n_{\alpha^{(m)}}}  \left(u^{(m)}(x_i, t_\ell) - \hat{u}^{(m)}(x_i, t_\ell) \right)^2 \label{eq:MSE_eq}.
\end{equation}
In theory, the loss can be averaged/batched across multiple trajectories $m$, however, in all experiments we use a single trajectory ($M=1$) during training. 
Parameters $\theta$ are optimized using standard gradient-based methods, where gradients are obtained by differentiating the solution map induced by the learned flow map and propagating sensitivities over time from the final observation back to earlier states. In practice, this is carried out by automatic differentiation through the discrete time-integration procedure used to construct the trajectory.

\subsection{Data generation, baselines, and training}\label{subsec:methods_datagen}
Solver settings, meshing, sampling, and preprocessing for the four examples are given in SI Appendix, section~S1. Architectures and hyperparameters of G-FuNK and of the two baselines, Geo-FNO \cite{li2023fourier} and a message-passing GNN \cite{iakovlev2020learning}, together with parameter counts and training settings, are given in SI Appendix, section~S2. All models were integrated with the same forward Euler neural ODE scheme and the same fixed time step within each example, and all were trained and tested on identical splits. Training was performed on a single NVIDIA A4500 GPU using \texttt{PyTorch}.

\section*{Acknowledgements}

We would like to thank the NIH for sponsoring this research (R01HL174440 and R01HL166759), and a Leducq Foundation grant.
MM was partially supported by AFOSR award FA9550-23-1-0445 and NSF award DMS-2436738.

\begin{itemize}
\item \textbf{Funding}  

This work was supported by NIH (R01HL174440 and R01HL166759), Leducq Foundation, AFOSR (FA9550-23-1-0445), and NSF (DMS-2436738).

\item \textbf{Conflict of interest/Competing interests}  

The authors declare no competing interests.

\item \textbf{Ethics approval and consent to participate}  

All patient-specific data used in this study were collected under protocols approved by the appropriate Institutional Review Board (IRB). All procedures complied with institutional guidelines and the Declaration of Helsinki. Informed consent was obtained from all participants where required.

\item \textbf{Data availability}

The non-patient benchmark datasets generated and analyzed in this study are publicly available.
The anisotropic heat equation dataset is available on
\href{https://doi.org/10.5281/zenodo.22771557}{Zenodo},
the reaction--diffusion dataset at
\href{https://doi.org/10.5281/zenodo.22667356}{Zenodo},
and the FitzHugh--Nagumo dataset at
\href{https://doi.org/10.5281/zenodo.22667186}{Zenodo}.
Owing to temporary difficulties uploading the complete datasets to Zenodo,
the corresponding files are also available through a
\href{https://livejohnshopkins-my.sharepoint.com/:f:/g/personal/sloeffl2_jh_edu/IgCI088okUSKRIaZxcjei6lrAfsys81RxMQisIpdjpSGT7k?e=4wQeBA}{public OneDrive repository}.
The Zenodo records will serve as the permanent archival repositories once
the remaining uploads are completed. Patient-specific cardiac geometries
and associated simulation data are not publicly available because of
institutional and participant privacy restrictions, but may be available
from the corresponding authors upon reasonable request and subject to
institutional approval.

\item \textbf{Code availability}

The source code for G-FuNK, together with scripts for training and evaluating the proposed model on the benchmark problems presented in this study, is publicly available at \url{https://gitlab.com/shaneloeffler/gfunk/-/tree/main}.

\item \textbf{Author contributions}  

SEL, ZA, DP, NT, MM conceived the study. SEL developed the methodology and performed the experiments. SEL, SC, SYA, CY, KS, AY, YL generated the patient-specific models and simulation data. NT, MM supervised the project. All authors contributed to manuscript preparation and approved the final manuscript.
\end{itemize}

\bibliographystyle{unsrtnat}%
\bibliography{sn-bibliography}%

\end{document}

\maketitle

\tableofcontents
\newpage

\section{Data generation}\label{sec:si_datagen}
In this section we provide the details for the data generation in all of the examples considered in this work.

\subsection{Data Generation for the Anisotropic Heat Example}
\label{heat_data_gen}
For this section of work, the heat equation was solved using the \texttt{NDSolve} function from the Mathematica software on the 2D unit square where the initial condition, $u_0$, was unique for each trajectory and chosen as a superposition of three Gaussian distributions where each center, variance, and covariance were picked independently from $\mathrm{Unif}([0,1])$, $\mathrm{Unif}([0.1,0.2])$, and $\mathrm{Unif}([-0.1,0.1])$, respectively. The primary direction of diffusion was defined as the normalized vector field $\hat{F}_{0,\beta}^{(1)}$, given by

\begin{equation}
\mathbf{F}_{0, \beta}^{(1)}(x,y) =
\begin{bmatrix}
\sin\left(a_1 \frac{x+a_2}{2 \pi}\right)
+\cos\left(a_3 \frac{y+a_4}{2 \pi}\right)
+a_5 x+a_7 y +c_a \\
\sin\left(b_1 \frac{x+b_2}{2 \pi}\right)
+\cos\left(b_3 \frac{y+b_4}{2 \pi}\right)
+b_5 x+b_7 y +c_b
\end{bmatrix},
\qquad
\hat{F}_{0,\beta}^{(1)}
=
\frac{\mathbf{F}_{0,\beta}^{(1)}}
{\left|\mathbf{F}_{0,\beta}^{(1)}\right|_2}.
\label{heat_fibers_eq_2}
\end{equation}

The second diffusion direction was defined to be the pointwise orthogonal complement of $\hat{F}_{0,\beta}^{(1)}$, such that

\begin{equation}
\hat{F}_{0,\beta}^{(2)}(x,y)
=
\begin{bmatrix}
-\hat{F}_{0,\beta,y}^{(1)}(x,y) \\
\hat{F}_{0,\beta,x}^{(1)}(x,y)
\end{bmatrix},
\label{heat_fibers_eq_3}
\end{equation}

ensuring that $\hat{F}_{0,\beta}^{(1)}$ and $\hat{F}_{0,\beta}^{(2)}$ form an orthonormal basis at each spatial location.

\begin{figure}[!htbp]
    \centering
\includegraphics[width=0.8\textwidth]{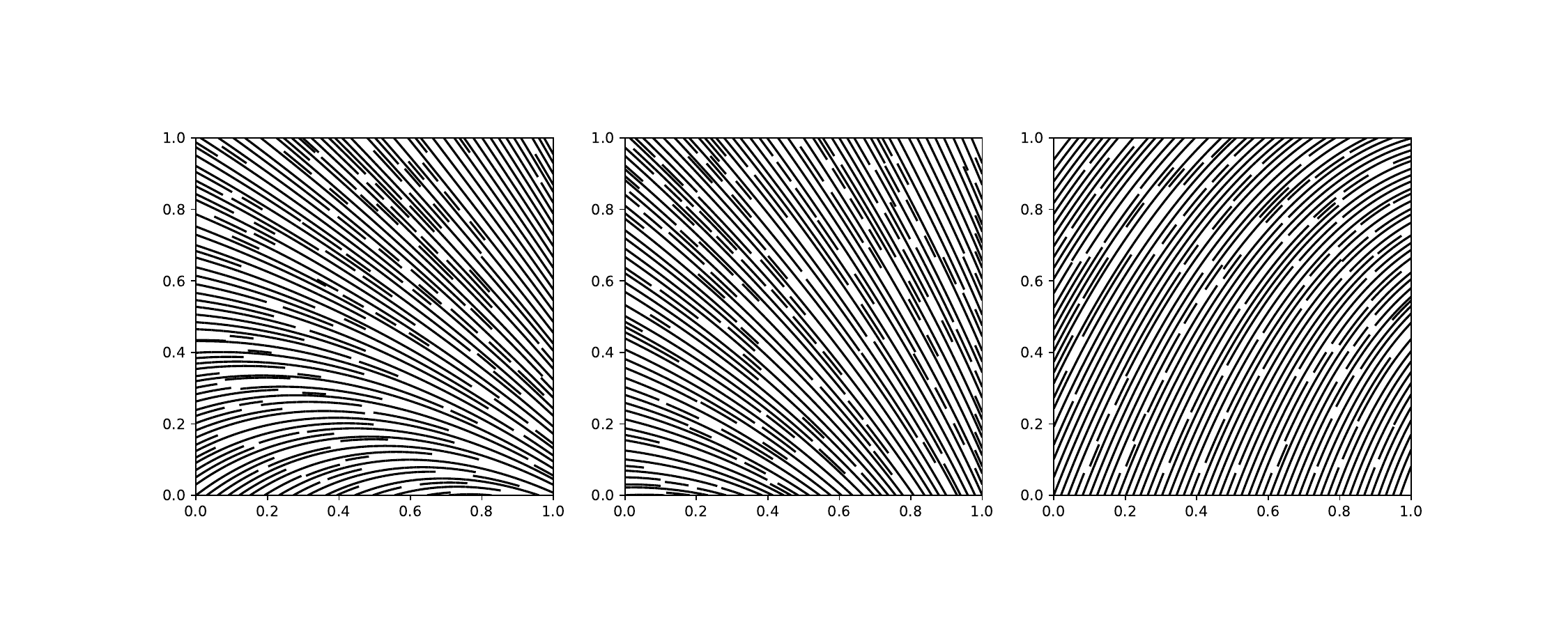} %
    \caption{Three different and unique examples of the primary diffusion vector fields used for the heat equation example in section~2.2.1 of the main text. Each example represents a diffusive field that was randomly determined for each trajectory.}
\label{heat_fibers_fig_1}
\end{figure}

For each $\beta$, the parameters $a_1,\ldots,a_5,a_7$ and $b_1,\ldots,b_5,b_7$ were sampled (independently) from the following distributions and were unique for each trajectory in the training and test data sets: $a_1,b_1,a_3,b_3\sim\mathrm{Unif}([0,2])$, $a_2,b_2,a_4,b_4\sim\mathrm{Unif}([-1,1])$,
$a_5,b_5,a_7,b_7\sim\mathrm{Unif}([-2,2])$.
The additive terms were defined as $c_a=a_6+a_8$ and $c_b=b_6+b_8$, where $a_6,a_8,b_6,b_8\sim\mathrm{Unif}([-1,1])$ independently. Thus, $c_a$ and $c_b$ follow the triangular distribution induced by the sum of two independent $\mathrm{Unif}([-1,1])$ random variables.
Representative diffusion-direction fields sampled in this manner are illustrated in Fig.~\ref{heat_fibers_fig_1} (heat equation results in section~2.2.1 of the main text).

After all simulations were completed, an unstructured triangular discretization of the unit-square domain was generated using Mathematica's \texttt{DiscretizeRegion}, with a maximum cell measure of $0.0002$. The resulting mesh contained 3952 discretized points. Linear interpolation was used to generate the trajectories at these points from $0$ to $0.02$ time units at intervals of $0.001$ time units. In addition, the diffusion tensor and vector defining the primary diffusion direction were evaluated at each discretized point for subsequent use. The training and test datasets consisted of $500$ and $100$ trajectories, respectively.

\subsection{Data Generation for the 2D Nonlinear Reaction--Diffusion Example}
\label{rd_data_gen}
For this section of work, the 2D Reaction--Diffusion equations were solved using the \texttt{openCARP} software which used a finite-element solver on a random rectangle to solve the Courtemanche equations. The solver used a Crank--Nicolson method with a time step of $\SI{5}{\micro\second}$ and an average edge length of $0.2$ mm. Each edge of the rectangle was drawn independently from $\mathrm{Unif}([15,30])~\mathrm{mm}$. The primary direction of diffusion was defined in the $x$-direction with a $5$ to $1$ anisotropic ratio and a base diffusion value of 0.0625 \nicefrac{mS}{mm}.

After all simulations were completed, a down-sampled point-cloud representation was constructed from the finite-element mesh and nearest-neighbor interpolation was used to generate the trajectories used for training at discretized points within the domain with a resolution of one point per $0.4$ mm voxel (via \texttt{open3d} \texttt{voxel\_down\_sample\_and\_trace}), and from $10$ to $100$ ms at $10$ ms intervals for a total of $90~\mathrm{ms}$. For consistency, time was subsequently shifted such that this snapshot defines $t_0=0$ in the surrogate dataset; thus, dataset time $t_0=0$ corresponds to $\SI{10}{\milli\second}$ in the original simulation. In addition, all other data, such as diffusion, vector orientation, etc., were calculated at each discretized point. For model training, the transmembrane potential was affinely normalized to $[0,1]$ using fixed bounds of $-85$ and $20~\mathrm{mV}$, such that $\tilde{u}=(u+85)/105$. Prior to evaluation, model predictions were transformed back to the original physical units, and all reported errors were calculated in these units. The same normalization was applied to all models. Training and test data sets consisted of $500$ and $50$ trajectories, respectively.

\subsection{Data Generation for the FitzHugh--Nagumo Example} \label{FN_data_gen}
In Eq.~5 of the main text, for the training and test sets, the parameters were chosen such that $A = 30$ and $\varepsilon = 0.2$ are fixed parameters. The cubic reaction term $f(u) = A\,u(u - \varepsilon)(1 - u)$ admits three spatially uniform equilibria at $u = 0$, $u = \varepsilon$, and $u = 1$, with $u = \varepsilon$ unstable. This structure supports bistable front propagation between the two stable states $u = 0$ and $u = 1$.

In this example, the \textit{effective} diffusion tensor $\mathbf{K}_{\alpha, \beta}(\mathbf{x})$ is a spatially varying diagonal tensor with a fixed anisotropy ratio of 2:1:
\begin{equation}
    \mathbf{K}_{\alpha, \beta}(\mathbf{x}) = \phi_{\alpha, \beta}(\mathbf{x}) \, \Sigma=  \phi_{\alpha, \beta}(\mathbf{x})\begin{bmatrix} 0.2 & 0 \\ 0 & 0.1 \end{bmatrix}\,,
    \label{FN_K}
\end{equation}
with the scalar diffusion profile $\phi$ given by
\begin{equation}
    \phi_{\alpha, \beta}(x,y) = 1 - \exp\!\left(-\frac{1}{5}\left(\frac{\tilde{x}^2}{r_x} + \frac{\tilde{y}^2}{r_y}\right)\right) + 0.02\,, \qquad
    \begin{cases} \tilde{x} = x\cos\theta - y\sin\theta \\ \tilde{y} = x\sin\theta + y\cos\theta \end{cases}\,,
    \label{FN_phi}
\end{equation}
with $r_x \sim \mathrm{Unif}([5,10])$, $r_y \sim \mathrm{Unif}([5,10])$, and $\theta \sim \mathrm{Unif}([0,\pi/2])$ drawn independently from each other and for each realization. The profile $\phi_{\alpha, \beta}$ attains its minimum value of $0.02$ at the origin and approaches $1.02$ away from it, so that the overall diffusion magnitude varies across the domain while the 2:1 directional ratio is preserved everywhere.

Each domain $\Omega_{\alpha}$ is a rectangle centered at the origin with side lengths $L_x \sim \mathrm{Unif}([5,10])$ and $L_y \sim \mathrm{Unif}([5,10])$, drawn independently, discretized on a uniform grid with spacing $\Delta x = \Delta y = 0.05$. Simulations were generated using an RK2 integrator with time step $\Delta t=\frac{\Delta x}{20}$, run for a total of 25 time units. The initial condition for each trajectory is a circular seed of unit amplitude centered at a randomly chosen interior point $\mathbf{x}_0 = (x_0, y_0)$:
\begin{equation*}
    u(\mathbf{x}, 0) = \mathbf{1}_{\left\{\,|\mathbf{x} - \mathbf{x}_0|^2 < 0.2\,\right\}}\,,
\end{equation*}
where $x_0 \sim \mathrm{Unif}\bigl([\,-L_x/2 + \sqrt{0.2},\; L_x/2 - \sqrt{0.2}\,]\bigr)$ and $y_0 \sim \mathrm{Unif}\bigl([\,-L_y/2 + \sqrt{0.2},\; L_y/2 - \sqrt{0.2}\,]\bigr)$, ensuring the support of $u(\cdot,0)$ is within the domain. 
This localized initial condition triggers a bistable front that propagates outward from $\mathbf{x}_0$, with the propagation speed and shape governed by the local structure of $\mathbf{K}_{\alpha, \beta}(\mathbf{x})$.
After the simulations were completed, the discrete sample points were determined using a Poisson sampling method with a minimum radius of $0.15$, and all data were linearly interpolated onto these points. For each trajectory, the initial condition was taken at the simulation time $t_0=2.5$ time units. The trajectory was then evolved through the original simulation time $t=22.5$ time units, corresponding to a total duration of $20$ time units sampled at $9$ temporal points with a spacing of $\Delta t'=2.5$ time units. A total of $510$ domains were generated, each with multiple independent trajectory simulations; $500$ domains with $1$ trajectory were used in training and $10$ were used in testing with $20$ trajectories each.

\subsection{Data Generation for the Cardiac Electrophysiology Example}
\label{CE_data_gen}

From cardiac CT scans of $25$ patients, the left atria were segmented, and a surface mesh was extracted using the \texttt{Seg3D} software. Manual corrections were made using \texttt{MeshMixer} and the average edge length was set to $0.25$ mm. Surfaces were then annotated for calculating universal atrial coordinates for mapping atrial fibers with an atlas-based method \cite{Ali2021,Roney2021,Roney2019}. The Courtemanche equations \cite{courtemanche1998ionic} were solved using the \texttt{openCARP} software, which used a finite-element solver that implemented a Crank--Nicolson scheme with a time step of $\SI{5}{\micro\second}$.

The diffusion tensor $\mathbf{K}_{\alpha, \beta}(\mathbf{x})$ was calculated using the following equations:

\begin{equation}
\mathbf{\hat{F}}_{\alpha, \beta}(\mathbf{x}) =
\left[
  \begin{array}{ccc}
    \vrule & \vrule & \vrule\\
    \hat{F}_{\alpha, \beta}^{(1)}(\mathbf{x}) & \hat{F}_{\alpha, \beta}^{(2)}(\mathbf{x}) & \hat{F}_{\alpha, \beta}^{(3)}(\mathbf{x}) \\
    \vrule & \vrule & \vrule 
  \end{array}
\right]
\end{equation}

\begin{equation}
\mathbf{K}_{\alpha, \beta}(\mathbf{x}) =
\mathbf{\hat{F}}_{\alpha, \beta}(\mathbf{x})
\left[
  \begin{array}{ccc}
    \sigma^{(1)} & 0 & 0\\
    0 & \sigma^{(2)} & 0 \\
    0 & 0 & \sigma^{(3)} 
  \end{array}
\right]
\mathbf{\hat{F}}_{\alpha, \beta}^T(\mathbf{x})\,,
\label{diffusiontensoreq} 
\end{equation}
where $\hat{F}_{\alpha, \beta}^{(1)}$ is the mapped fiber field from the atlas-based method described in \cite{Roney2021,Ali2021, Roney2019}, $\hat{F}_{\alpha, \beta}^{(1)}$, $\hat{F}_{\alpha, \beta}^{(2)}$, and $\hat{F}_{\alpha, \beta}^{(3)}$ are orthonormal, and $\sigma^{(2)}=\sigma^{(3)}=0.625$ \nicefrac{mS}{mm}, and $\sigma^{(1)}=5\sigma^{(2)}$.

After all simulations were completed, a down-sampled mesh was computed using the software \texttt{Open3D} in which the original finite-element mesh was downsampled using the function \nolinkurl{simplify_quadric_decimation} with a target number of triangles of $10{,}000$, resulting in meshes with approximately $5000$ vertices. Nearest-neighbor interpolation was used to map the time-evolving transmembrane potential, diffusion, and fiber orientations onto the downsampled mesh. To exclude the initial stimulus transient, surrogate training trajectories were initialized from the simulation snapshot at $\SI{10}{\milli\second}$ after stimulation. For consistency, time was subsequently shifted such that this snapshot defines $t_0=0$ in the surrogate dataset; thus, dataset time $t_0=0$ corresponds to $\SI{10}{\milli\second}$ in the original simulation. The trajectories were then down-sampled to a temporal resolution of $\SI{5}{\milli\second}$ over the resulting $0$--$\SI{85}{\milli\second}$ dataset-time window. For model training, the transmembrane potential was affinely normalized to $[0,1]$ using fixed bounds of $-85$ and $20~\mathrm{mV}$, such that $\tilde{u}=(u+85)/105$. Prior to evaluation, model predictions were transformed back to the original physical units, and all reported errors were calculated in these units. The same normalization was applied to all models. Each domain consisted of $20$ trajectories, for a total of $500$ trajectories used across the training/test splits.

\section{Network descriptions and training}
\label{sub:network_descriptions}

\paragraph{Training Parameters.}
All models presented in this work used a time-averaged mean squared error loss function as described in Eq.~18 of the main text. The parameters for all models were updated using an Adam optimizer with a step learning rate scheduler that reduced the learning rate by half every 50 epochs. For the 2D nonlinear reaction--diffusion example, the scheduler interval was instead set to 25 epochs for G-FuNK and the GNN and 100 epochs for Geo-FNO.

For each benchmark, initial learning rates were 0.001, 0.0001, 0.0001, and 0.0001 for the heat, 2D nonlinear reaction--diffusion, FitzHugh--Nagumo, and cardiac electrophysiology examples, respectively. For optimization stability, the initial learning rate was adjusted to 0.0005 for the GNN on the heat equation example, 0.00005 for Geo-FNO and the GNN on the 2D nonlinear reaction--diffusion example, and 0.0003 for Geo-FNO on the FitzHugh--Nagumo example.

Training was terminated when the training loss failed to improve for 50 consecutive epochs, with a maximum of 500 epochs; the checkpoint with the lowest training loss was retained. No separate validation set was used; the test set was held out exclusively for final evaluation. All models were trained with a batch size of 1 trajectory. To improve training stability, gradient norm clipping was applied for models exhibiting unstable optimization. G-FuNK used clipping thresholds of 1.0 for the heat equation and 0.5 for the remaining examples (2D nonlinear reaction--diffusion, FitzHugh--Nagumo, and cardiac electrophysiology). Among the baselines, Geo-FNO used a threshold of 0.5 for both the 2D nonlinear reaction--diffusion and FitzHugh--Nagumo, and GNN used thresholds of 1.0, 1.0, and 0.5 for the heat equation, 2D nonlinear reaction--diffusion, and FitzHugh--Nagumo examples, respectively. Training and test sets contained distinct domain and parameter configurations and all models were trained with identical train/test splits for a fair comparison. Training was performed on a single NVIDIA A4500 GPU using \texttt{PyTorch}. During training for the cardiac electrophysiology example, trajectories were divided into three sub-trajectories to reduce memory requirements, while reported errors were computed over the full trajectories.

\paragraph{G-FuNK Network Description.} \label{sec:gfunk_network_description}
For all four examples, the same G-FuNK network architecture was used. The model consists of: (1) a linear projection to a latent dimension $d_p$ from an input dimension $d_{\text{input}}=1$; (2) four G-FuNK layers, each followed by a GeLU activation; and (3) a multilayer perceptron (MLP) that maps from the latent dimension $d_p$ to an intermediate dimension of $d_{\text{int}}$ (32 in all examples), followed by a GeLU activation and a final linear projection to a single output dimension of $d_{\text{out}} = 1$.
Within each G-FuNK layer, the latent functions defined on the graph nodes are projected onto the $k_{\max}$ lowest graph Fourier modes, where $k_{\max}$ controls the bias of the model in terms of how well it can approximate the solution operator in the spectral domain, and is set here to $k_{\max}=100$ ($k_{\max}=50$ for the anisotropic heat and 2D nonlinear reaction--diffusion examples). Spectral multipliers are learned using multi-head self-attention applied to the scalar eigenvalues associated with these modes. Each of the $k_{\max}$ scalar eigenvalues is independently projected from dimension one into an embedding space of dimension $d_{\text{SA}}$ (we used $d_{\text{SA}}=64$ in all examples), producing the query, key, and value representations for four attention heads. The embedded eigenvalue representations are then used to compute self-attention across the $k_{\max}$ eigenvalues and are subsequently mapped from $d_{\text{SA}}$ dimensions back to a single scalar for each eigenvalue. These scalar spectral multipliers are applied to each Fourier mode, and the result is projected back to the graph domain using the inverse graph Fourier transform (IGFT). This construction requires only $7d_{\text{SA}}+1$ trainable parameters per self-attention block, while its computational cost is $\mathcal{O}(k_{\max}^2d_{\text{SA}}+k_{\max}d_{\text{SA}}^2)$ due to the attention operation. We note that the lack of an FFT is not problematic here, since only a modest number, $k_{\max}$, of Fourier modes is used, so the complexity of the GFT and IGFT is a modest $O(n_\alpha k_{\max})$.

\begin{table}[htbp]
    \centering
    \small
    \renewcommand{\arraystretch}{1.2}
    \setlength{\tabcolsep}{3pt}
    \begin{tabularx}{\linewidth}{@{}>{\raggedright\arraybackslash}p{2.8cm}>{\raggedright\arraybackslash}p{3.1cm}rr>{\raggedright\arraybackslash}X@{}}
    \toprule
    \textbf{Section} & \textbf{Parameter formula} & \textbf{Ex. 1, 3} & \textbf{Ex. 2, EP} & \textbf{Computational cost} \\
    \midrule
    (1) Input projection & $2d_p$ & 400 & 300 & $\mathcal{O}(n_\alpha d_u d_p)$ \\
    \addlinespace
    \multicolumn{5}{@{}l}{(2) Spectral branch} \\
    Fourier transform (and inverse) & & & & $\mathcal{O}(n_\alpha k_{\max}d_p)$ \\
    Self-attention blocks (\texttt{R0--R3}) & $7d_{\text{SA}}+1$ & 1796 & 1796 & $\begin{aligned}[t]\mathcal{O}(&k_{\max}^2d_{\text{SA}}\\&{}+k_{\max}d_{\text{SA}}^2)\end{aligned}$ \\
    \addlinespace
    \multicolumn{5}{@{}l}{(2) Spatial branch} \\
    Linear projections (\texttt{W0--W3}) & $d_p(d_p+1)$ & 160800 & 90600 & $\mathcal{O}(n_\alpha d_p^2)$ \\
    \addlinespace
    (3) Output MLP & $\begin{aligned}[t]&d_{\text{int}}(d_p+d_{\text{out}})\\&{}+d_{\text{int}}+d_{\text{out}}\end{aligned}$ & 6465 & 4865 & $\mathcal{O}(n_\alpha d_p d_{\text{out}})$ \\
    \midrule
    \textbf{Total} & & \textbf{169461} & \textbf{97561} & \\
    \bottomrule
    \end{tabularx}
    \caption{Parameter counts and computational costs at prediction time for G-FuNK in examples 1, 2, 3 and cardiac electrophysiology (EP). In the spectral and spatial branches, formulas are per block/layer, while the numerical parameter counts include all four G-FuNK layers.}
    \label{tab:param_comparison}
\end{table}

The output of this spectral branch is aggregated with a linear transformation of the latent node features and passed through a GeLU activation.
For the heat and FitzHugh--Nagumo examples, $d_p$ was set to 200. For the 2D reaction--diffusion and cardiac electrophysiology examples, $d_p$ was set to 150. 
The computational cost, per layer, at prediction time, scales linearly in the number of spatial points $n_\alpha$, quadratically in both the latent dimension $d_p$ and the spectral truncation $k_{\max}$. 
The details can be found in Table \ref{tab:param_comparison}.

For all four examples, a symmetric normalized graph Laplacian was constructed. Edge weights were defined using the Gaussian kernel (Eq.~13 of the main text) for the heat, 2D nonlinear reaction--diffusion, and FitzHugh--Nagumo examples, and using a geodesic distance--based scheme for the cardiac electrophysiology case. Connectivity was determined via radius-based neighborhoods, with radii of $0.25$, $\SI{5}{\milli\meter}$, and $5$ for the respective examples, and a geodesic radius of $\SI{20}{\milli\meter}$ for the cardiac case. The full construction of the edge weights is described in section~4.1.1 of the main text.

\paragraph{Geo-FNO/FNO Network Description.}
Geo-FNO models were implemented by adapting the publicly available implementation from the elasticity example provided with the original Geo-FNO study. The 2D Geo-FNO architectures consist of: (a) a linear lifting layer that maps the input features---comprising the solution $u$, spatial coordinates $(x,y)$, and diffusion information---to a latent representation; (b) a sequence of Fourier layers with GeLU activations; and (c) a projection head consisting of two linear layers, expanding to a width of 128 and then mapping to the scalar output. For the 2D nonlinear reaction--diffusion example, five Fourier layers with 10 Fourier modes in each spatial direction, a latent width of 32, and a $40\times40$ computational Fourier grid were used. For the FitzHugh--Nagumo example, five Fourier layers with 10 modes in each spatial direction, a latent width of 40, and a $32\times32$ computational Fourier grid were used. For the anisotropic heat equation, four Fourier layers with 16 Fourier modes in each spatial direction, a latent width of 40, and a $32\times32$ computational Fourier grid were used.

The Geo-FNO model takes as input the current state $u$, spatial coordinates $(x,y)$, and problem-specific diffusion information. Spatial coordinates were first scaled to a common reference coordinate system and translated such that the center of each sample's bounding box was located at \((0.5,0.5)\). For the varying-domain reaction--diffusion and FitzHugh--Nagumo examples, the transformed coordinates are subsequently passed through an IPHI deformation network adapted from the elasticity example of the original Geo-FNO implementation. The IPHI network takes the transformed spatial coordinates together with their radial distance and angular position relative to the center of the reference domain. These features are augmented with sinusoidal and cosine positional encodings and passed through a multilayer perceptron, which outputs a two-dimensional coordinate deformation that is applied multiplicatively to the input coordinates before construction of the non-uniform Fourier representation. The learned deformation is omitted for the anisotropic heat equation because the spatial domain is fixed across samples. Diffusion information for all examples was represented by the flattened diffusion tensor components at each query point.

\paragraph{GNN Network Description.}
For the GNN baseline, we employ a message-passing neural network similar to \cite{iakovlev2020learning}, consisting of separate message and aggregation networks. The input features are the current state \(u\), spatial coordinates \((x,y)\), and the flattened diffusion tensor components at each query point. For the 2D random rectangle, the primary diffusion direction is instead provided at each query point.

The message network computes pairwise interactions between node \(i\) and its neighbors \(j\). Its inputs include the state at node \(i\), the difference in state between nodes \(i\) and \(j\), their relative positions, the diffusion tensor at node \(i\), and the difference in diffusion features between the two nodes. The latter feature is omitted for the 2D nonlinear reaction--diffusion example, for which the diffusion is constant. These inputs are processed by a five-layer multilayer perceptron with width 60 and hyperbolic tangent activations, followed by a final linear layer with output width 40.

The aggregation network combines the aggregated messages at each node with its local features. Aggregation was implemented using the mean function. Specifically, the aggregated message, the state at node \(i\), and the diffusion features at node \(i\) are concatenated and passed through a five-layer MLP with width 60 and hyperbolic tangent activations, followed by a final linear projection to a scalar output. For the heat example, the position was also passed to the aggregation network since the domain was fixed across all examples.

For the GNN baseline, graph connectivity was defined using radius-based neighborhoods, with smaller neighborhood radii than those used for G-FuNK due to memory constraints associated with message-passing on dense graphs. The radii were set to \(0.05\), \(\SI{1}{\milli\meter}\), and \(0.35\) for the heat, 2D nonlinear reaction--diffusion, and FitzHugh--Nagumo examples, respectively.

\paragraph{Neural ODE Parameters.}
All models were integrated in time using the same neural ODE framework and numerical scheme to ensure a fair comparison across methods.
In all experiments, a forward Euler scheme was employed with a fixed time step which was kept consistent across models within each example.
For the heat, 2D nonlinear reaction--diffusion, FitzHugh--Nagumo, and cardiac electrophysiology examples, the corresponding fixed time steps were $0.0001$, $\SI{1}{\milli\second}$, $0.2$, and $\SI{0.2}{\milli\second}$, respectively. 
During training of the models for the cardiac electrophysiology example, the trajectories were divided into three sub-trajectories to reduce memory requirements, with each sub-trajectory initialized independently from its corresponding ground-truth state. Thus, errors do not accumulate between sub-trajectories during training. At test time, the model was rolled out over the full original trajectories without resetting to the ground truth, and errors are reported over these full trajectories in the held-out geometry test split.

\section{Additional Results}

\begin{figure}[!htbp]
    \centering

    \begin{subfigure}{0.8\textwidth}
        \centering
        \includegraphics[trim={10 5 8 5},clip,width=.78\textwidth]{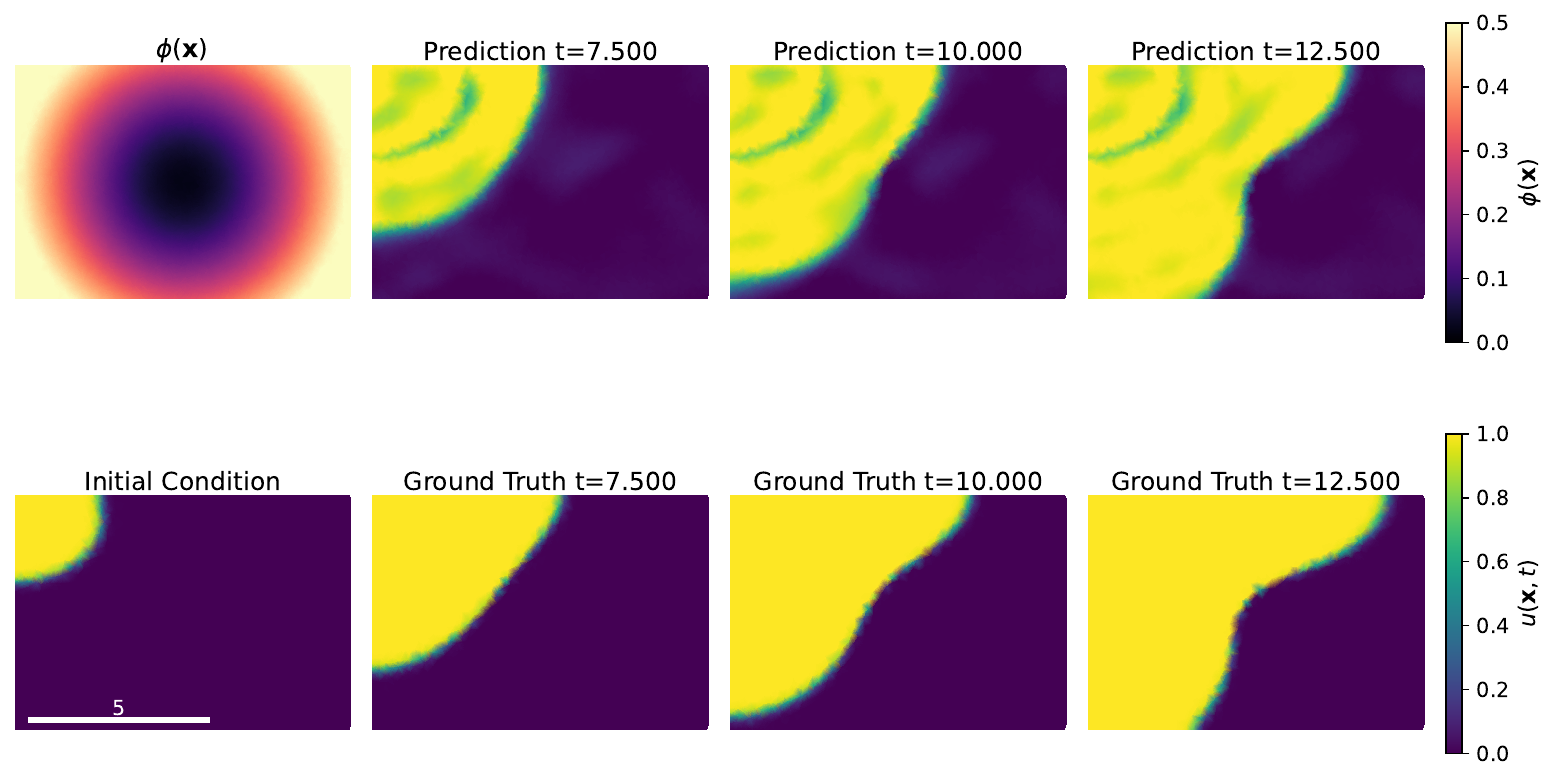}
        \caption{}
        \label{fig:geofno_fnA}
    \end{subfigure}

    \vspace{0.25em}
    \hrule
    \vspace{0.25em}

    \begin{subfigure}{0.8\textwidth}
        \centering
        \includegraphics[trim={10 5 8 5},clip,width=.78\textwidth]{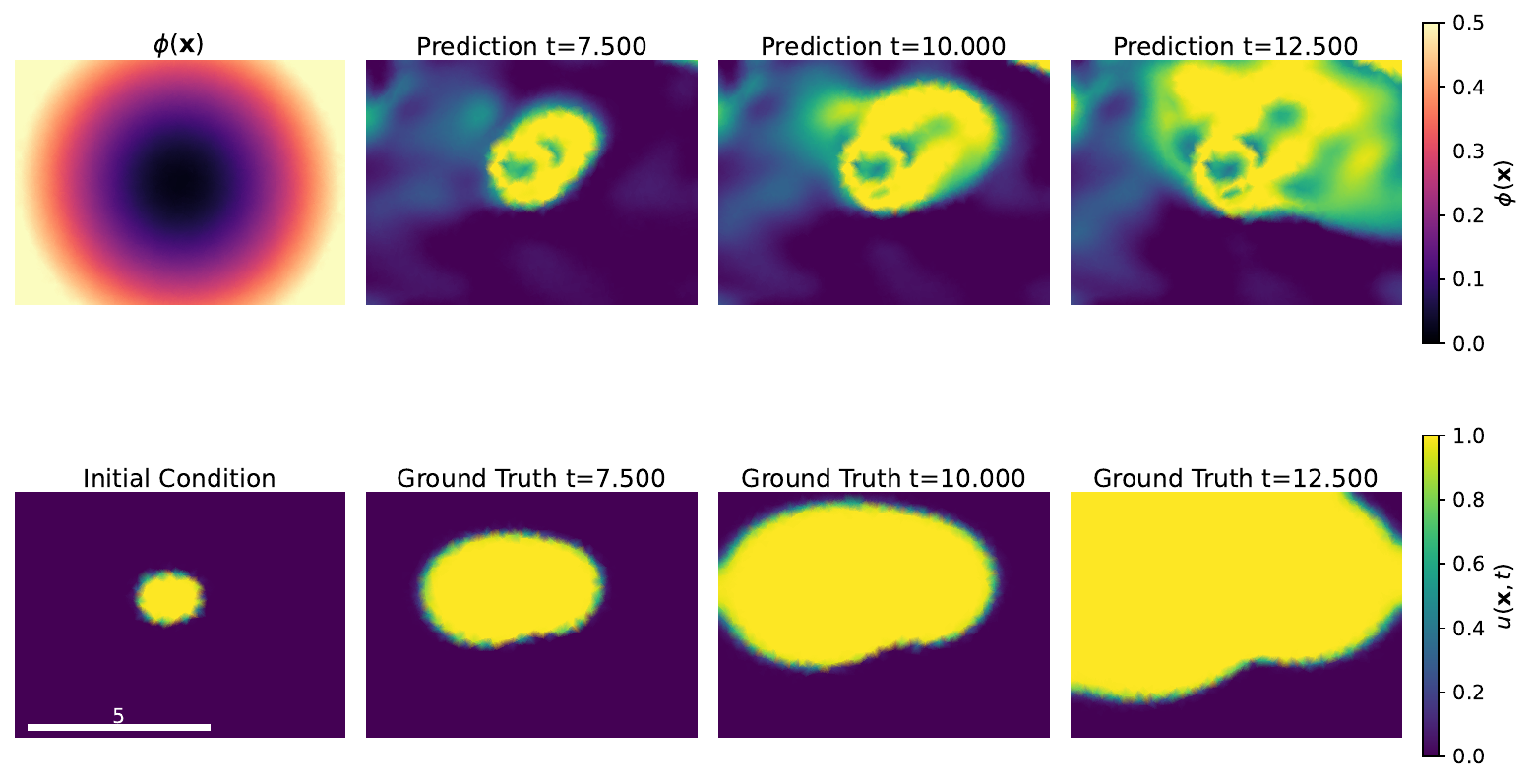}
        \caption{}
        \label{fig:geofno_fnB}
    \end{subfigure}

    \vspace{0.25em}
    \hrule
    \vspace{0.25em}

    \begin{subfigure}{0.8\textwidth}
        \centering
        \includegraphics[trim={10 5 8 5},clip,width=.78\textwidth]{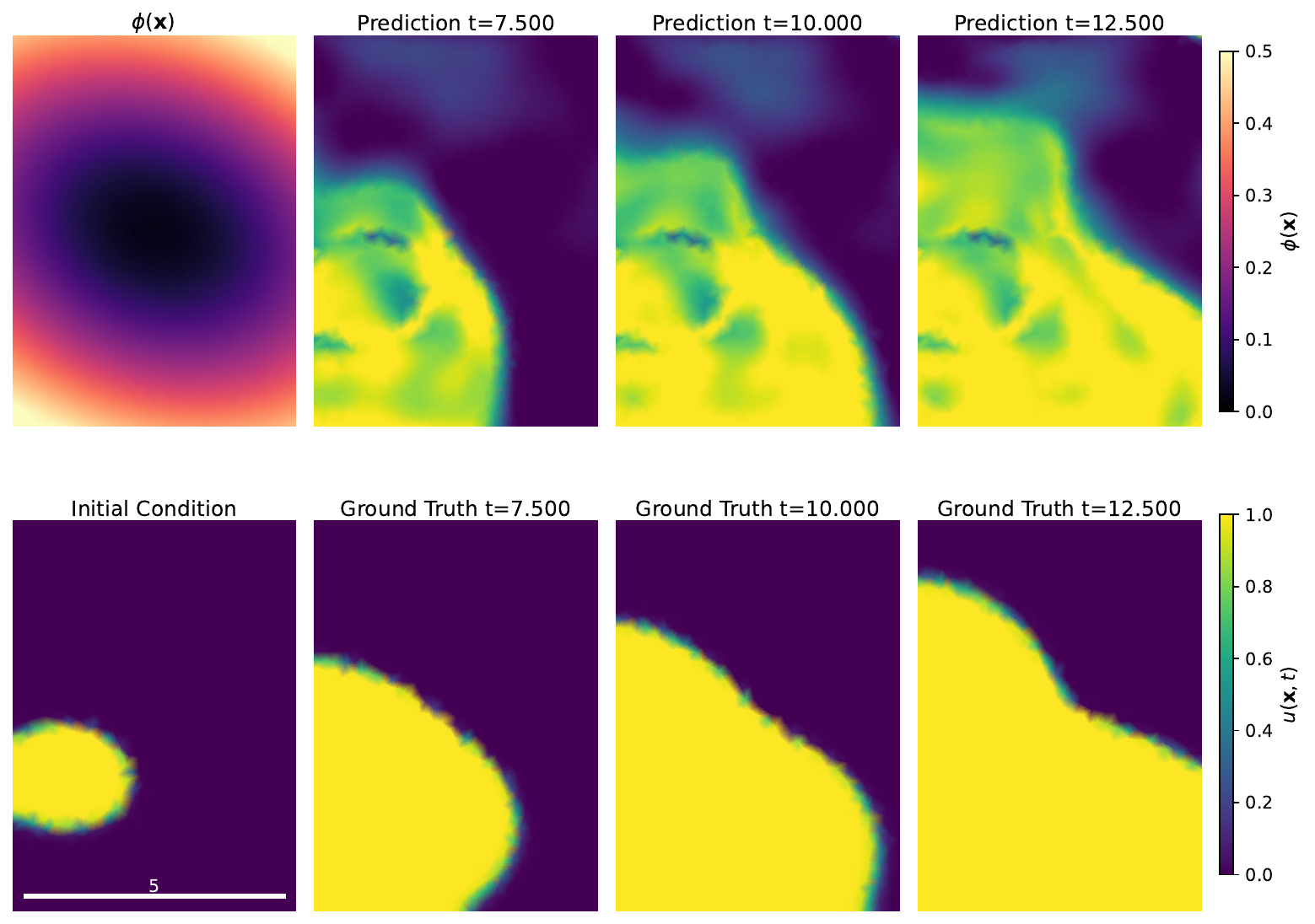}
        \caption{}
        \label{fig:geofno_fnC}
    \end{subfigure}

    \caption{\textbf{FitzHugh--Nagumo with Various Domain and Diffusive Parameters: Geo-FNO vs Target (Three Test Examples).}
        Three representative test trajectories corresponding to different initial conditions, different Gaussian-defined anisotropic diffusive fields and domains. In each example, the \textit{top-left} panel shows the pointwise multiplying function $\phi_{\alpha, \beta}(\mathbf{x})$ that defines the local strength of the anisotropic (ratio 2:1) diffusive field $\mathbf{K}_{\alpha, \beta}(\mathbf{x})$, as defined in Eq.~5 of the main text. The \textit{bottom-left} panel shows the initial condition at $t_0=2.5$. The remaining panels show snapshots of the solution at three later time points, with the Geo-FNO predictions shown above the corresponding ground truth targets. Panel (a) and panel (b) are the same domain and parameter set but different initial conditions. Panel (c) shows a domain and scaling factor that are more skewed compared with panels (a) and (b). 
        }
    \label{fig:geo-fno-FN}
\end{figure}

Figure~\ref{fig:geo-fno-FN} shows representative Geo-FNO predictions on the FitzHugh--Nagumo benchmark of the main text. Their accuracy varies with domain size, diffusivity field, and initial condition.

\begin{figure*}[htbp]
\centering

\begin{minipage}{0.48\textwidth}
    \centering
    \includegraphics[
        width=\linewidth,
        trim={0 0 0 0.75cm},
        clip
    ]{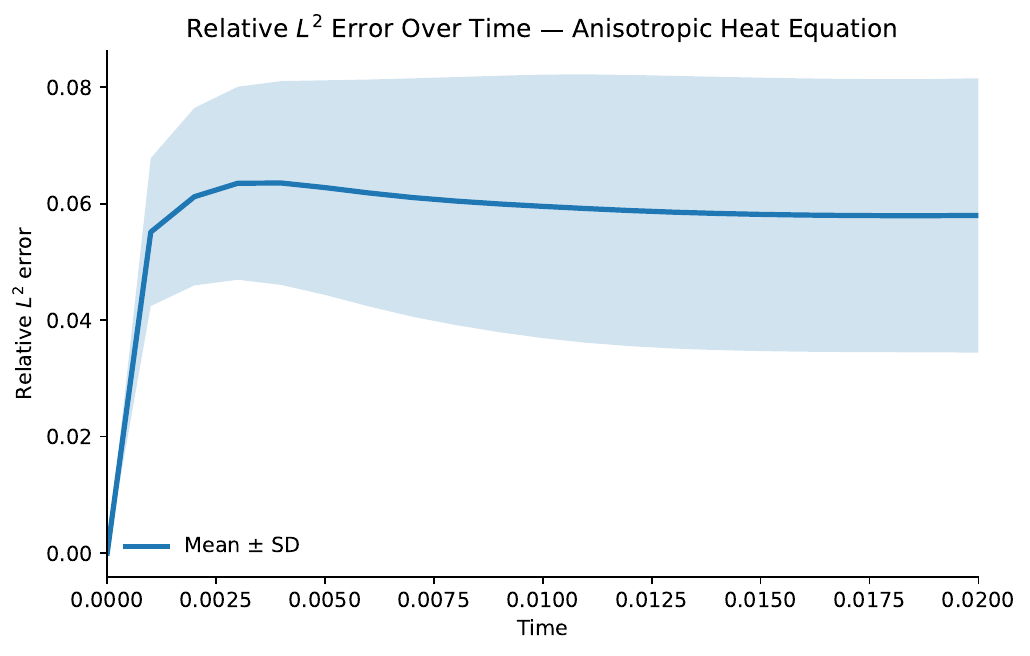}

    \small (a) Anisotropic Heat
\end{minipage}
\hfill
\begin{minipage}{0.48\textwidth}
    \centering
    \includegraphics[
        width=\linewidth,
        trim={0 0 0 0.75cm},
        clip
    ]{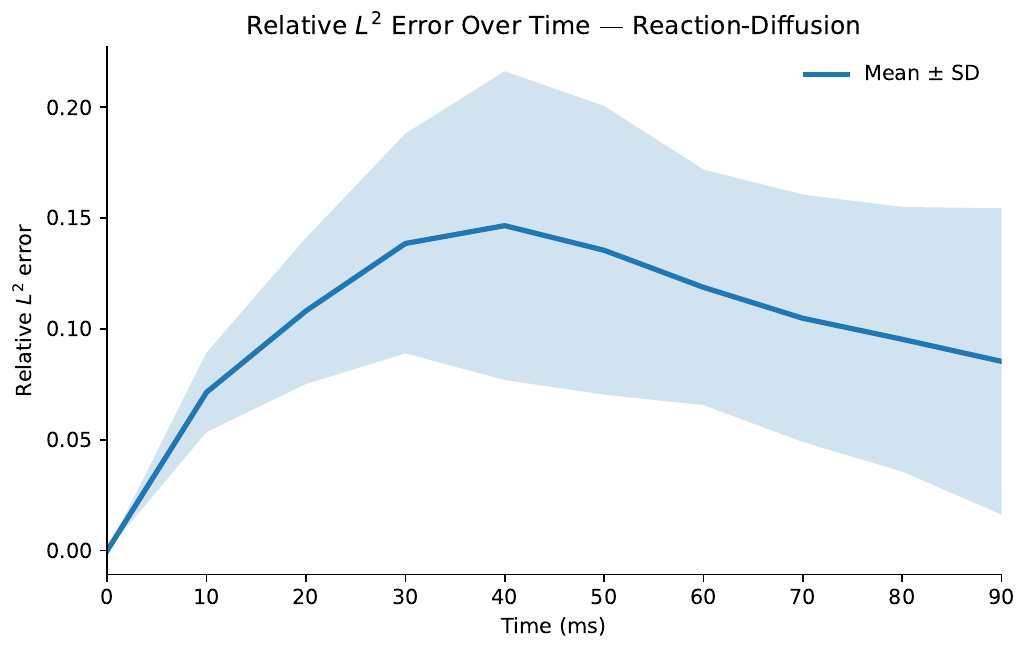}

    \small (b) Reaction--Diffusion
\end{minipage}

\vspace{0.5em}

\begin{minipage}{0.48\textwidth}
    \centering
    \includegraphics[
        width=\linewidth,
        trim={0 0 0 0.75cm},
        clip
    ]{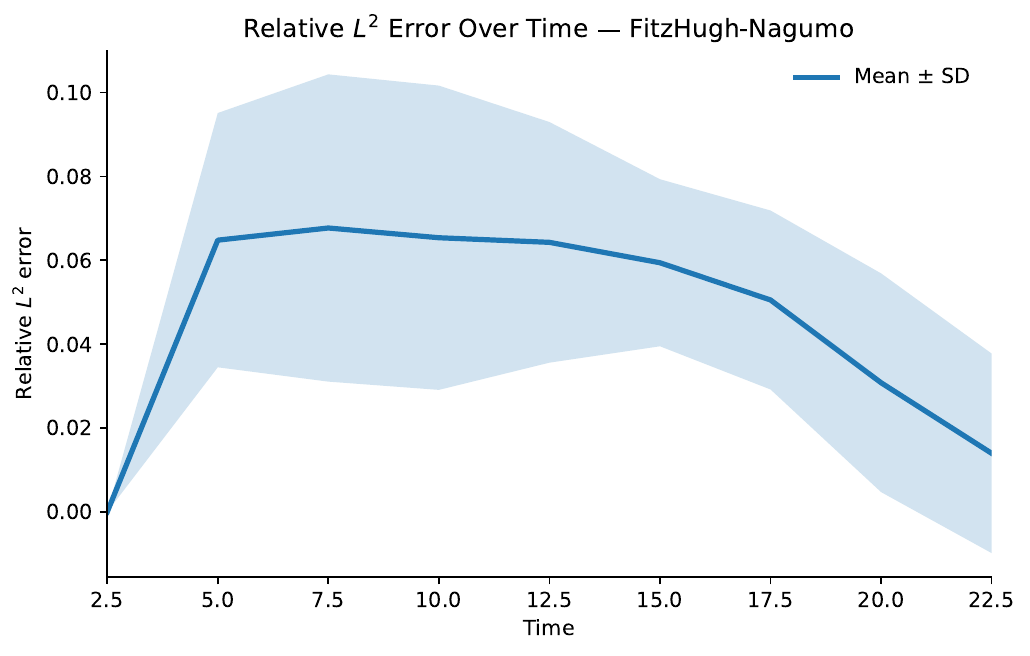}

    \small (c) FitzHugh--Nagumo
\end{minipage}
\hfill
\begin{minipage}{0.48\textwidth}
    \centering
    \includegraphics[
        width=\linewidth,
        trim={0 0 0 0.75cm},
        clip
    ]{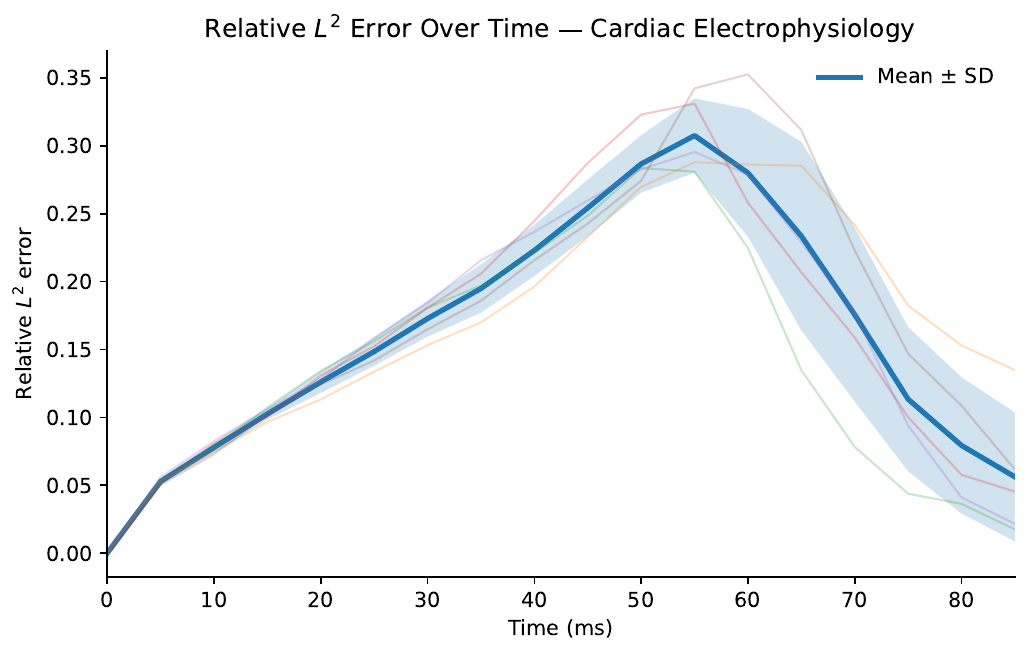}

    \small (d) Cardiac electrophysiology
\end{minipage}

\caption{\textbf{Temporal evolution of relative $L^2$ error.}
Time-dependent relative $L^2$ errors for G-FuNK across the four benchmark
problems. Solid lines indicate the mean across test trajectories and shaded
regions indicate $\pm$ one standard deviation. For cardiac electrophysiology,
statistics are computed across the five single-geometry hold-out evaluations.}
\label{fig:temporal_error}
\end{figure*}

The time-dependent relative $L^2$ errors for the four benchmarks of the main text are shown in Fig.~\ref{fig:temporal_error}. The relatively large variability observed at some time points reflects the diversity of the test trajectories and domains, which span different initial conditions, geometries, and parameter configurations. For propagating solutions, this variability is further amplified by the sensitivity of the relative $L^2$ error to small differences in wavefront position, such that trajectories with similar qualitative dynamics can exhibit substantially different instantaneous errors.

\begin{figure}[!htbp]
    \centering
    \includegraphics[trim={0 0 0 0},clip,width=.8\textwidth]{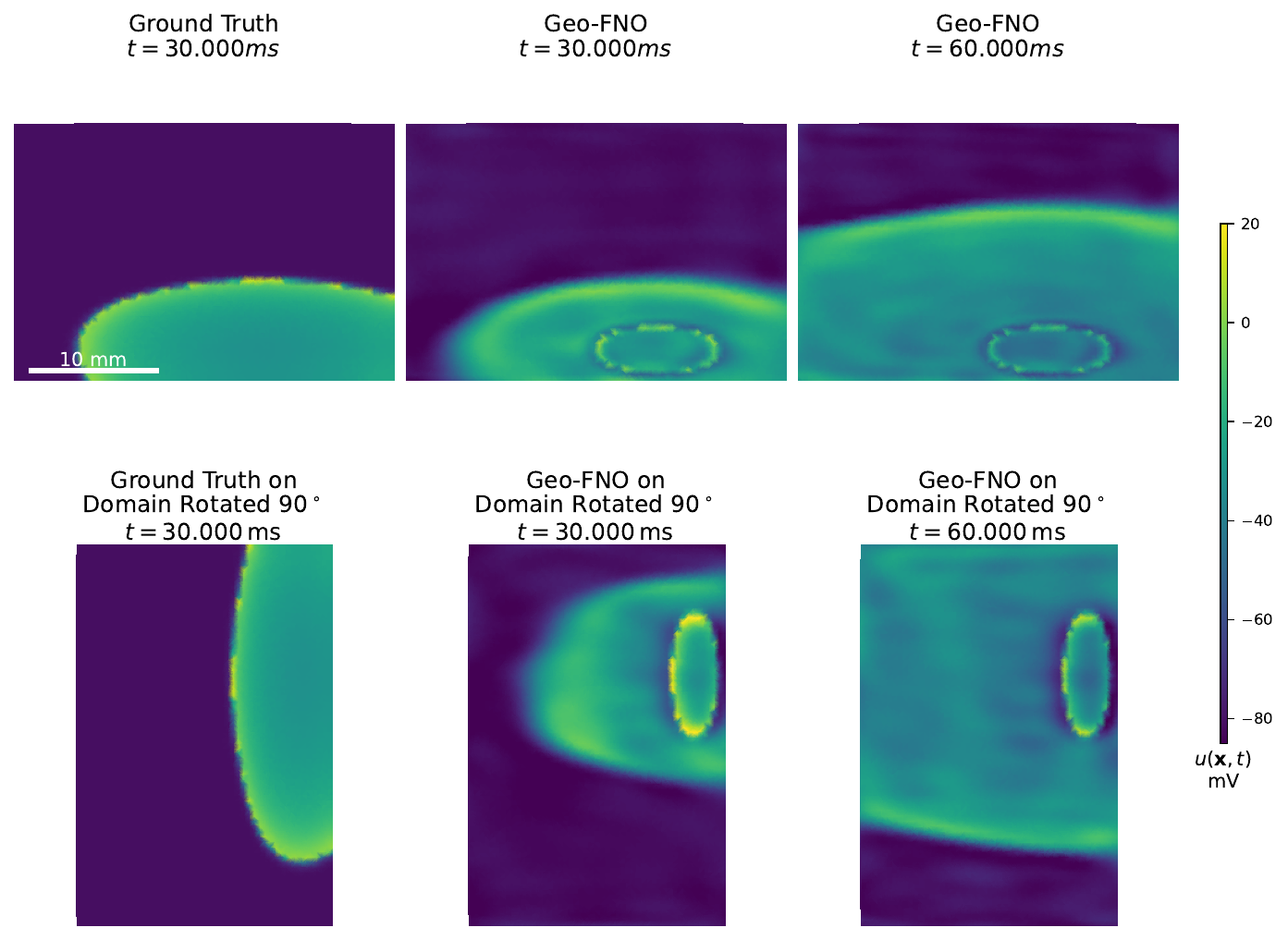}

    \caption{Geo-FNO Performance: Target vs Prediction for the random rectangle case. \textbf{Top} Geo-FNO performance on the original test set. \textbf{Bottom} Geo-FNO performance on the same test set but rotated $90^\circ$. Times are measured from the initial condition at $t=0$. When the domain is rotated $90^\circ$ the performance of Geo-FNO decreased as the rel. $L^2$ error increased from $0.112 \pm 0.015$ to $0.450 \pm 0.039$. }
    \label{GeoFNO_rand_rec_fig}
\end{figure}

The performance of Geo-FNO/FNO is intrinsically tied to the underlying coordinate system as observed in Figure \ref{GeoFNO_rand_rec_fig}. While Geo-FNO performed quite well on the test set, the underlying assumptions of the network are rooted in a Euclidean coordinate system and the model was not generalizable to simple rotations. We note that domain matching would resolve these issues, but this simple toy example is used to highlight potential issues when using embeddings intrinsically tied to a global coordinate system. 

\section{Graph-construction sensitivity experiments}

\subsection{G-FuNK}
\paragraph{Laplacian effect on G-FuNK performance.}
We investigate the sensitivity of G-FuNK to the choice of graph Laplacian by evaluating several graph weighting and normalization strategies. Because the graph Laplacian defines the spectral basis used throughout G-FuNK, its construction directly influences the learned representation. This study evaluates the robustness of the framework to reasonable choices of graph constructions. The main manuscript reports results obtained using the Gaussian kernel-weighted symmetric normalized graph Laplacian. Here, we compare this choice with alternative graph constructions and discuss their empirical performance and training stability.

We consider two edge-weighting schemes: the Gaussian kernel weighting used throughout the main text and an edge-conductivity weighting. These are defined as

\begin{align}
    \text{Gaussian: }\, W_{ij} &=\frac{1}{D_{\text{weight}}}\exp\!\left(-\frac{(x_j-x_i)^T\mathbf{K}_{ij}^{-1}(x_j-x_i)}{2\sigma^2}\right) \\
    \text{Edge-Conductivity: }\,W_{ij} &=\frac{(x_j-x_i)^T \cdot \overline{\mathbf{K}}_{ij} \cdot (x_j-x_i)}{\lVert x_j-x_i\rVert^2}\,
\end{align}

where $\mathbf{K}^{-1}_{ij}=\frac12 \left( \mathbf{K}^{-1}(x_i)+\mathbf{K}^{-1}(x_j) \right)$, $D_{\text{weight}} = \left(\det(\mathbf{K}(x_i))\det(\mathbf{K}(x_j))\right)^{1/4}$, and $\overline{\mathbf{K}}_{ij}=\frac12 \left( \mathbf{K}(x_i)+\mathbf{K}(x_j) \right)$ with $x$ as the position of the nodes $i$ or $j$.

We further investigated the effect of graph Laplacian normalization by considering the symmetric normalized and the random-walk normalized graph Laplacians, defined respectively as

\begin{align*}
    \mathcal{L}_{\text{Sym}} &= I-D^{-1/2} W D^{-1/2} \\
    \mathcal{L}_{\text{RW}} &= I- D^{-1}W \\
\end{align*}

where $D$ and $W$ are the degree and weight matrices. Because the random-walk Laplacian is generally non-symmetric, its eigenvectors are orthogonal with respect to the degree-weighted inner product rather than the standard Euclidean inner product. Consequently, the graph Fourier transform becomes $\hat{a} = \mathcal{F}_{\text{RW}}a := (\langle \psi_i, a \rangle_D)_{i=1,\dots,{n_{\alpha}}}=\Psi^TD a$ where $D$ is the degree matrix; the inverse Fourier transform under the metric $D$ has the same form as what is presented in the main text $a = \mathcal{F}^{-1}_{\text{RW}}\hat{a} := \sum_{i=1}^{n_{\alpha}} \hat a(\lambda_i)\psi_i = \Psi \hat{a}$. 

Table \ref{tab:laplacian_normalization} summarizes the performance of G-FuNK for each graph construction and normalization strategy.

\begin{table}[ht]
    \centering
    \begin{tabular}{ccc>{\raggedright\arraybackslash}p{3.8cm}}
        \toprule
        \multicolumn{4}{c}{Anisotropic Heat - Parameter Adaptation} \\
        \midrule
        Weighting & Normalization & Rel. $L^2$ Error & Additional Notes \\
        \midrule
        Gaussian & Symmetric & 0.056 & Radius = 0.25 \\
        Gaussian & Symmetric & 0.070 & Radius = 0.2 \\
        Edge-Conductivity & Symmetric & 0.032 & Radius = 0.25 \\
        \midrule
        \multicolumn{4}{c}{FitzHugh--Nagumo} \\
        \midrule
        Weighting & Normalization & Rel. $L^2$ Error & Additional Notes \\
        \midrule
        Gaussian & Symmetric & 0.046 & Radius = 5, $dt=0.2$ \\ 
        Gaussian & Symmetric & 0.044 & Radius = 5, $dt=0.05$ \\
        Gaussian & Random-Walk & 0.077 & Radius = 5, $dt=0.05$ \\
        Edge-Conductivity & Symmetric &  0.153 & Radius = 1.5, $dt=0.05$ \\
        Edge-Conductivity & Random-Walk & 0.179 & Radius = 1.5, $dt=0.05$ \\
        \midrule
        \multicolumn{4}{c}{Cardiac Electrophysiology - Five geometry-level hold-out splits} \\
        \midrule
        Weighting & Normalization & Rel. $L^2$ Error & Additional Notes \\
        \midrule
        Gaussian & Symmetric & 0.160 & Radius = 20 mm \\
        Edge-Conductivity & Symmetric & 0.171 & Radius = 10 mm \\
        Gaussian & Random-Walk & 0.193 & Radius = 10 mm, a single hold-out split \\
        \bottomrule
    \end{tabular}
    \caption{Comparison of graph construction and Laplacian normalization strategies across benchmark problems. The first configuration listed for each example corresponds to the model reported in the main text. $dt$ refers to the time step used for training the model.}
    \label{tab:laplacian_normalization}
\end{table}

\section{Resolution Transferability}
\label{sec:resolution_transfer}

To evaluate the ability of G-FuNK to generalize across graph resolutions, a cardiac G-FuNK model was trained using approximately 5,000 graph nodes and evaluated without retraining on the same held-out geometries represented at different resolutions. This analysis was performed on a single hold-out split from the five single-geometry hold-out evaluations and is intended as a targeted assessment of resolution transfer rather than an additional cross-validation result. For each resolution, the reference finite-element trajectories were independently interpolated from the original finite-element meshes onto the corresponding graph using nearest-neighbor interpolation. Relative $L^2$ errors were then computed between the predicted and reference trajectories at each resolution.

As shown in Table~\ref{tab:resolution_transfer}, G-FuNK maintained comparable accuracy over a range of graph resolutions surrounding the training resolution. The relative $L^2$ error was 0.255 at approximately 2,500 nodes, 0.187 at approximately 4,000 nodes, 0.173 at the training resolution of approximately 5,000 nodes, 0.163 at approximately 6,000 nodes, and 0.165 at approximately 10,000 nodes. The approximately 5,000-node result corresponds to the baseline G-FuNK error for this hold-out split rather than the mean across the five hold-out splits. The model therefore transferred to both coarser and finer graph resolutions without retraining, including evaluation at approximately twice the training resolution. The degradation observed at approximately 2,500 nodes indicates that sufficiently dense spatial sampling remains important for accurately representing the underlying graph operator, while performance remained stable across the approximately 4,000--10,000 node range.

\begin{table}[!htbp]
\centering
\caption{Resolution transferability of G-FuNK.
Relative $L^2$ error of a cardiac G-FuNK model trained at approximately 5,000 nodes and evaluated without retraining at different graph resolutions on the same held-out geometries.}
\label{tab:resolution_transfer}
\begin{tabular}{cc}
\toprule
Number of nodes & Relative $L^2$ error \\
\midrule
$\sim$2,500  & 0.255 \\
$\sim$4,000  & 0.187 \\
$\sim$5,000  & 0.173 \\
$\sim$6,000  & 0.163 \\
$\sim$10,000 & 0.165 \\
\bottomrule
\end{tabular}
\end{table}

\section{Training-Set Size Convergence}
\label{sec:training_size}

To assess how G-FuNK accuracy depends on the amount of training data, we repeated the anisotropic heat experiment while reducing the number of training trajectories (each associated with a unique diffusion field and initial condition) from 500 to 250, 100, and 50. All other settings were fixed across runs; only the number of training trajectories changed. Each model was evaluated on the same 100 held-out test trajectories at 21 time points.

As shown in Table~\ref{tab:training_size}, the rel. $L^2$ error decreased steadily with the number of training trajectories, from $0.1201$ with 50 to $0.0734$ with 250, a reduction of about $39\%$. The 500-trajectory model reported in the main text reaches $0.0564$. For the three reduced-training models, the standard error is approximately $\mathrm{std}/\sqrt{100}$ ($0.0025$--$0.0037$), so the differences between successive training-set sizes exceed the sampling uncertainty of the test set.

\begin{table}[!htbp]
\centering
\caption{\textbf{Convergence with training-set size for the anisotropic heat equation.} Rel. $L^2$ error of G-FuNK on 100 held-out test trajectories (21 time points each) as a function of the number of training trajectories. Std.\ is the standard deviation across test trajectories. The 500-trajectory value is the model reported in the main text.}
\label{tab:training_size}
\begin{tabular}{ccc}
\toprule
Training trajectories & Rel. $L^2$ error & Std. \\
\midrule
50  & 0.120 & 0.037 \\
100 & 0.099 & 0.030 \\
250 & 0.073 & 0.025 \\
500 & 0.056 & 0.018  \\
\bottomrule
\end{tabular}
\end{table}

\section{Sensitivity to Temporal Resolution and Integration Scheme}

We evaluated the learned FitzHugh--Nagumo and cardiac electrophysiology generators using Forward Euler and fourth-order Runge--Kutta (RK4) across a range of inference temporal step sizes. For each example, the inference step size \(dt'\) was varied from \(4\) to \(\frac{1}{4}\) times the corresponding training step size. For the FitzHugh--Nagumo example, models trained using Forward Euler with \(dt=0.2\) and \(dt=0.05\) time units were evaluated, while the cardiac electrophysiology generator was trained using Forward Euler with \(dt=0.2\,\mathrm{ms}\).

The results demonstrate that the learned dynamics exhibit sensitivity to both the temporal resolution and integration scheme used during inference. For the FitzHugh--Nagumo model trained with \(dt=0.2\), the lowest error was obtained when inference was performed using the same Forward Euler step size used during training. Reducing the inference step size did not improve accuracy and instead produced progressively larger errors, indicating that the learned generator is partially coupled to the temporal discretization used during training. In contrast, the model trained with the finer \(dt=0.05\) step exhibited relatively stable performance for inference step sizes at or below the training step size.

For both FitzHugh--Nagumo models and the cardiac electrophysiology model, increasing the Forward Euler step size beyond that used during training led to increased error, with particularly large errors at \(dt'=4dt\). RK4 substantially reduced this instability at the coarsest inference step sizes, but did not consistently improve accuracy when the inference step size was equal to or smaller than that used during training. These results indicate that the learned generators can be integrated using alternative numerical schemes, but that the highest accuracy is generally obtained when the temporal discretization used during inference is consistent with that used during training. The cardiac electrophysiology sensitivity analysis was performed on a single hold-out split from the five geometry-level hold-out evaluations. The results are shown in Table~\ref{tab:time_integration_sen}.

\begin{table}[!htbp]
\centering
\caption{\textbf{Sensitivity to inference temporal resolution and integration scheme.}
Relative \(L^2\) errors for the learned FitzHugh--Nagumo and cardiac electrophysiology generators across inference step sizes and integration schemes. The FitzHugh--Nagumo G-FuNK model was trained with both \(dt=0.2\) and \(dt=0.05\) time units; the sensitivity analysis for the \(dt=0.05\) model was evaluated on 50 test trajectories. The cardiac electrophysiology G-FuNK model was trained with \(dt=0.2\,\mathrm{ms}\). All models were trained using Forward Euler, and inference was performed using Forward Euler and fourth-order Runge--Kutta (RK4).}
\label{tab:time_integration_sen}

\begin{tabular}{c|c|cc}
    \hline
    Benchmark & Inference \(dt'/dt\) & Forward Euler & RK4 \\
    \hline
    FHN & $4$ & 7.995 & 0.595 \\
    Trained with $dt=0.2$  & $2$ & 0.1292 & 0.171 \\
        & $1$ & 0.046 & 0.153 \\
        & $\frac12$ & 0.103 & 0.161 \\
        & $\frac14$ & 0.133 & 0.161 \\
    \hline
    FHN & $4$ & 1.968 & 0.301 \\
      Trained with $dt=0.05$  & $2$ & 0.181 & 0.077 \\
        & $1$ & 0.055 & 0.059 \\
        & $\frac12$ & 0.059 & 0.060 \\
        & $\frac14$ & 0.060 & 0.060 \\
    \hline
    Cardiac EP & $4$ & 0.456 & 0.237 \\
        & $2$ & 0.201 & 0.188 \\
        & $1$ & 0.173 & 0.188 \\
        & $\frac12$ & 0.177 & 0.188 \\
        & $\frac14$ & 0.182 & 0.188 \\
    \hline
\end{tabular}
\end{table}

\clearpage
\section{Full temporal rollout}

\begin{figure}[H]
    \centering
    \begin{subfigure}[t]{0.48\textwidth}
        \centering
        \includegraphics[width=\textwidth]{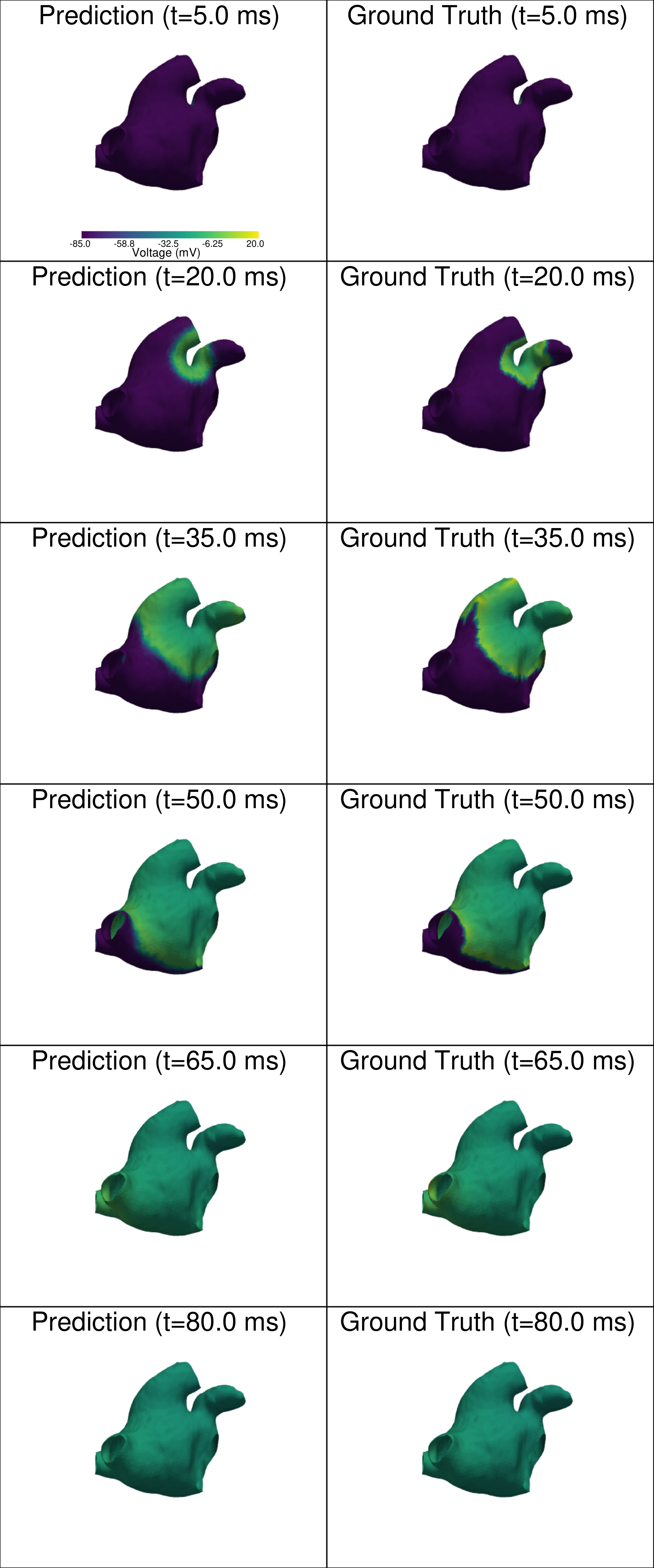}
        \caption{Septal Anterior View.}
        \label{fig:trajectory_0}
    \end{subfigure}
    \hfill
    \begin{subfigure}[t]{0.48\textwidth}
        \centering
        \includegraphics[width=\textwidth]{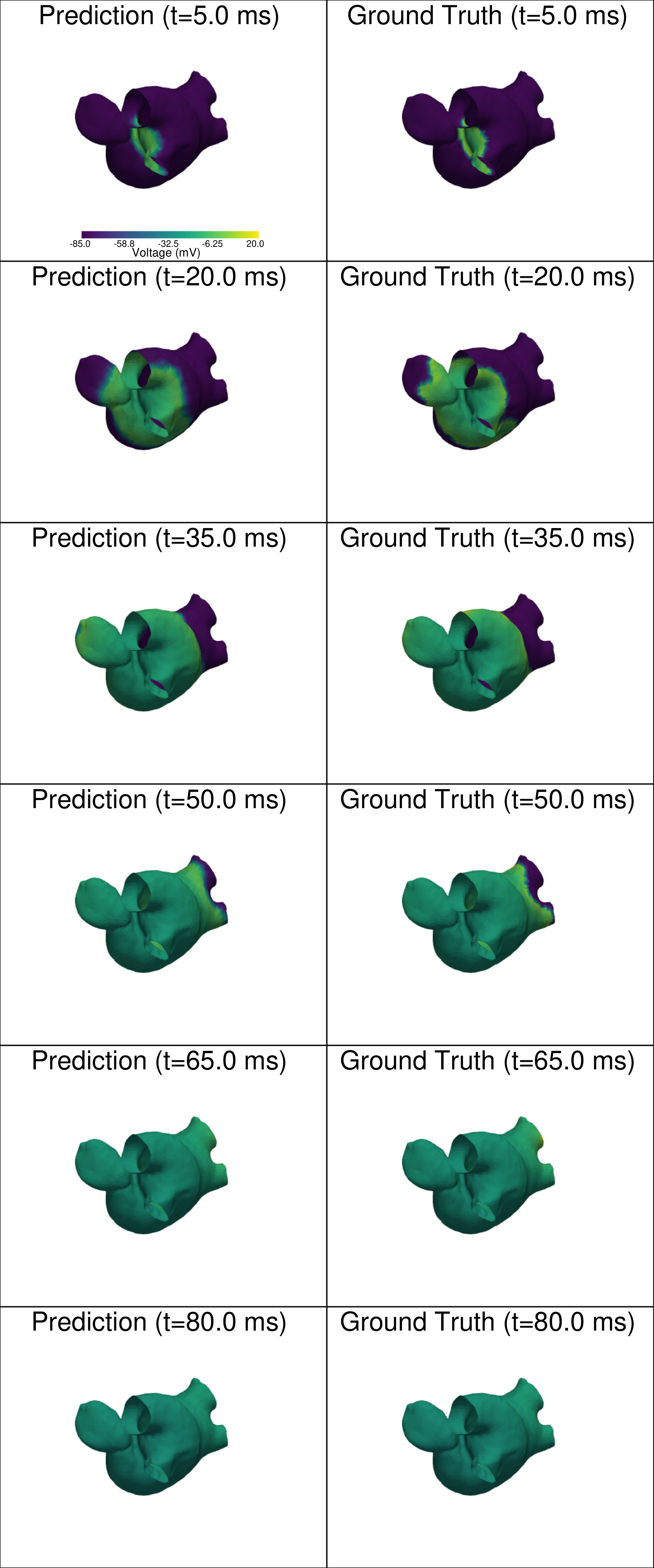}
        \caption{Posterior Roof View.}
        \label{fig:trajectory_90}
    \end{subfigure}

    \caption{Full-trajectory G-FuNK predictions versus ground truth from two different views.}
    \label{fig:full_trajectory}
\end{figure}

\FloatBarrier
\section{Operator-learning formulation}\label{sec:si_formulation}

\noindent \textbf{Problem setup}.
In this context, we focus on the family of semilinear differential operators, governing the evolution of a function \( u: \Omega_\alpha\times[0,\infty)\rightarrow\mathbb{R} \), $\mathbf{x}\in\Omega_\alpha$, $t\ge0$, in the form:
\begin{equation}
\left\{
\begin{aligned}
    \partial_t u(\mathbf{x},t) &= \nabla_{\mathbf{x}}\cdot
    (\mathbf{K}_{\alpha, \beta}(\mathbf{x}) \nabla_{\mathbf{x}} u(\mathbf{x},t))+\mathcal{S}(\,u(\mathbf{x}, t), \mathbf{x}, \nabla_\mathbf{x} u(\mathbf{x},t)\,) \\
    &=: \mathcal{N}(\, u(\mathbf{x}, t), \mathbf{K}_{\alpha, \beta}(\mathbf{x}), \mathbf{x}, t\,), 
    \\ 
    u(\mathbf{x},0) &= u_0(\mathbf{x}) \\
    0&= \left( \mathbf{K}_{\alpha, \beta}(\mathbf{x}) \nabla u(\mathbf{x}) \right) \cdot \mathbf{n}(\mathbf{x}) \:\,,\quad \forall \, \mathbf{x} \: \in \partial \Omega_\alpha\,
    \end{aligned}
    \right.
\end{equation}
where $\mathbf{K}_{\alpha, \beta}$ is a diffusion tensor field on the domain satisfying uniform ellipticity, and \( \mathcal{N} \) denotes the (true) generator of the dynamics.
For each fixed domain $\Omega_\alpha$, the evolution equation Eq.~1 of the main text defines a dynamical system on the function space $\mathcal{U}(\Omega_\alpha;\mathbb{R})$ consisting of the spatial component of $u$ at time $t$; such evolution can be written as
\begin{equation}
\frac{d}{dt} u(\cdot,t) = \mathcal{N}(u(\cdot,t); \mathbf{K}_{\alpha, \beta}, \Omega_\alpha).
\end{equation}
By time integration, this induces a flow map, taking an initial condition to the solution at time $t$:
\begin{equation}
\Phi_t^{\alpha, \beta} : \mathcal{U}(\Omega_\alpha;\mathbb{R}) \rightarrow \mathcal{U}(\Omega_\alpha;\mathbb{R}), 
\quad u_0 \mapsto u(\cdot,t),
\end{equation}

\noindent \textbf{Neural Operators for PDEs}.
In scientific machine learning, data-driven deep learning methods aim at predicting the solutions of parametrized families of partial differential equations (PDEs) faster than classical numerical integration methods used for large-scale simulations, typically by allowing large time steps, at the cost of some loss of accuracy. Neural operators aim at learning mappings between function spaces that generalize across a family of PDEs without retraining for varying parameters or (initial/boundary/...) conditions \cite{kovachki2023neural}. 
In this paper, we consider a neural operator as   $\mathcal{A}_{\theta}$ which is simply a neural architecture that represents a map
\begin{equation}  
\mathcal{A}_{\theta}: \mathcal{U}(\Omega_{\alpha};\mathbb{R}^{d_u}) \times \mathcal{M} \rightarrow \mathcal{U}(\Omega_{\alpha};\mathbb{R}^{d_v}),
\end{equation}
where $\mathcal{M}$ denotes the parameter space of pairs $(\Omega_\alpha,\mathbf{K}_{\alpha, \beta})$, which approximates a target solution operator $\mathcal{A}$. In the present setting, the solution operator can be written as
\begin{equation}
\mathcal{A}(u_0; \mathbf{K}_{\alpha, \beta}, \Omega_\alpha) := \Phi_T^{\alpha, \beta}(u_0),
\end{equation}
mapping an initial condition to the solution at a fixed time $T$.

\noindent \textbf{Learning the Generator and the Flow Map}.
Rather than learning the solution operator $\mathcal{A}$ directly, our objective is to learn the generator of the dynamics. Specifically, we learn a neural operator \( \mathcal{N}_{\theta} \) that approximates the generator $\mathcal{N}$ by mapping a current state $u(\cdot,t)$, together with parameters $\mathcal{G}_{\alpha, \beta} \in \mathcal{M}$, to an approximation of its time derivative.

Solutions are then obtained by integrating the learned generator. For a time step $\Delta t > 0$, we define a one-step update operator
\begin{equation}
\Psi_{\theta}^{\Delta t}(u;\mathcal{G}_{\alpha, \beta})
:=
u + \Delta t\,\mathcal{N}_{\theta}(u;\mathcal{G}_{\alpha, \beta}),
\end{equation}
so that
\begin{equation}
u(\cdot,t+\Delta t) \approx \Psi_{\theta}^{\Delta t}(u(\cdot,t);\mathcal{G}_{\alpha, \beta}).
\label{nn_approx}
\end{equation}

For $t/\Delta t\in\mathbb{N}$, this induces a learned flow defined by iterative composition:
\begin{equation}
\tilde{\Phi}_t^{\theta}(u_0;\mathcal{G}_{\alpha, \beta})
:=
\underbrace{\Psi_{\theta}^{\Delta t}\circ \cdots \circ \Psi_{\theta}^{\Delta t}}_{t/\Delta t \text{ times}}(u_0),
\end{equation}
which approximates the true flow $\Phi_t^{\alpha, \beta}$. More generally, $\Psi_{\theta}^{\Delta t}$ may be replaced by a higher-order numerical integrator (e.g., a neural ODE solver), yielding
\begin{equation}
\tilde{\Phi}_t^{\theta}(u_0)
:=
\mathcal{I}_t\big(\mathcal{N}_{\theta}; u_0, \mathcal{G}_{\alpha, \beta}\big).
\end{equation}

\bibliographystyle{unsrtnat}
\bibliography{sn-bibliography}